\documentclass[runningheads]{llncs}

\usepackage{accv}

\usepackage{graphicx}
\usepackage{booktabs}
\usepackage[ragged]{sidecap}    

\usepackage[accsupp]{axessibility}  % Improves PDF readability for those with disabilities.

\usepackage[dvipsnames]{xcolor}
\usepackage{multirow}

\newcommand\norm[1]{\left\lVert#1\right\rVert}
\DeclareMathOperator*{\argmax}{argmax}
\DeclareMathOperator*{\argmin}{argmin}

\usepackage[pagebackref,breaklinks,colorlinks,citecolor=accvblue]{hyperref}
\usepackage{orcidlink}

\usepackage{todonotes}

\begin{document}

% ---------------------------------------------------------------
% TODO REVIEW: Replace with your title
%\title{On Partial Shape Correspondence and Functional Maps} 
% \title{Partial shape correspondence revisited} 
\title{On Unsupervised Partial Shape Correspondence}

% TODO REVIEW: If the paper title is too long for the running head, you can set
% an abbreviated paper title here. If not, comment out.
% \titlerunning{On Partial Shape Correspondence and Functional Maps}
\titlerunning{On Unsupervised Partial Shape Correspondence}

% TODO FINAL: Replace with your author list. 
% Include the authors' OCRID for the camera-ready version, if at all possible.
\author{Amit Bracha \orcidlink{0009-0004-9496-5555}
\and 
Thomas Dag\`es \orcidlink{0000-0002-0803-9300} 
\and 
Ron Kimmel \orcidlink{0000-0002-3180-7961}
}

% TODO FINAL: Replace with an abbreviated list of authors.
% \authorrunning{F.~Author et al.}
\authorrunning{A.~Bracha et al.}
% First names are abbreviated in the running head.
% If there are more than two authors, 'et al.' is used.

% TODO FINAL: Replace with your institution list.
% \institute{Princeton University, Princeton NJ 08544, USA \and
% Springer Heidelberg, Tiergartenstr.~17, 69121 Heidelberg, Germany
% \email{lncs@springer.com}\\
% \url{http://www.springer.com/gp/computer-science/lncs} \and
% ABC Institute, Rupert-Karls-University Heidelberg, Heidelberg, Germany\\
% \email{\{abc,lncs\}@uni-heidelberg.de}}
\institute{Technion - Israel Institute of Technology\\
Haifa, Israel\\
\email{amit.bracha@cs.technion.ac.il}\\
{\vspace{-18em} 
\color{gray} \small Accepted for publication at the Asian Conference on Computer Vision (ACCV) 2024.\\
\vspace{1em}
This is an updated version of the arXiv preprint\\
``On Partial Shape Correspondence and Functional Maps'' arXiv:2310.14692 (2023).
}
\vspace{18em}
}

\maketitle

\begin{abstract}
    While dealing with matching shapes to their parts, we often apply a tool known as functional maps. 
    The idea is to translate the shape matching problem into ``convenient'' spaces by which matching is performed algebraically by solving a least squares problem. 
    Here, we argue that such formulations, though popular in this field, introduce errors in the estimated match when partiality is invoked. 
    Such errors are unavoidable even  for advanced feature extraction networks, and they can be shown to escalate with increasing degrees of shape partiality, adversely affecting the learning capability of such systems. 
    To circumvent these limitations, we propose a novel approach for partial shape matching. 
    
    Our study of functional maps led us to a novel method that establishes direct correspondence between partial and full shapes through feature matching bypassing the need for functional map intermediate spaces. 
    The Gromov Distance between metric spaces leads to the construction of the first part of our loss functions. 
    For regularization we use two options: a term based on the area preserving property of the mapping, and a relaxed version that avoids the need to resort to functional maps.
    
    The proposed approach shows superior performance on the SHREC'16 dataset, outperforming existing unsupervised methods for partial shape matching.
    Notably, it achieves state-of-the-art results on the SHREC'16 HOLES benchmark, superior also compared to supervised methods. 
    We demonstrate the benefits of the proposed unsupervised method when applied to a new dataset PFAUST for part-to-full shape correspondence. %at  \url{https://github.com/ABracha/PFAUST}. 

   % {\color{blue} 
    \keywords{Shape correspondence \and partial shapes \and learning to match}
    %}
\end{abstract}

\section{Introduction}
\label{sec:intro}

In recent years, the emergence of deep learning allowed significant leap forwards in efficiently solving computational problems, like finding the point-to-point correspondence between two sampled surfaces. 
And yet, for unsupervised partial shape matching the contribution of deep learning was somewhat limited.
Most state-of-the-art shape correspondence  approaches rely on the {\em functional maps} framework \cite{fm}, which searches for the mapping between spectral representations of corresponding features on the two input shapes. 
One popular framework uses a {\em functional map layer} \cite{litany2017deep} as a differentiable component in a neural network architecture.
We will show that such a {\em functional map layer} introduces inherent errors in the case of partial shape matching. 

Theoretically, we could overcome such unavoidable errors of using functional maps as layers in a neural network.
For example, one could first detect the corresponding parts by matching the spectra of an appropriate Hamiltonian of the part with that of the Laplacian of the whole
\cite{rampini2019correspondence, bensaid2023partialhamiltonian}.
After detecting the matching parts,
the point-to-point correspondence can be solved by resorting to an algebraic problem defined by the relation of projected scalar feature functions onto spectral domains, aka {\it functional maps}.

State-of-the-art shape correspondence approaches often involve three components \cite{halimi2019unsupervised,roufosse2019unsupervised,sharp2022diffusionnet,cao2023unsupervised,sharma2020weakly,geomfmnet}.
Initially, a network computes refined pointwise features.
Next, a {\em functional map layer} uses these features to estimate the  correspondence between the shapes.
Finally, loss functions based on the resulting functional map are applied during training to further adapt the features to the data at hand. 

In this paper, we analyse  the use of least squares for estimating the functional map in the {\em functional map layer} and focus on the resulting errors involved in partiality.
While previous studies argue that it should be possible to use least squares in the {\em functional map layer} to obtain the ground-truth map without errors \cite{DPFM}, here, we prove that the projected errors cannot be avoided in such a setting. 
Furthermore, the magnitude of the error directly relates to the degree of partiality -- the smaller the part of the shape the larger the error. 
The reason for this error stems from the fact that the least squares method considers the inner product of the shape's basis functions and the feature descriptors.
That inner product involves integration over the whole shape. 

With this observation in mind, we propose a novel approach for partial shape matching, surpassing all existing unsupervised techniques. 
The proposed method deviates from the conventional process of generating a functional map from the extracted features prior to the correspondence. 
We avoid the functional map formulation altogether and establish correspondence between partial and full shapes through direct feature matching.
In essence, we demonstrate that for partial shape matching, choosing an appropriate loss, like the Gromov Distance \cite{bronstein2006generalized,bronstein2010gromov,memoli2005distance}, is as important as operating in convenient spaces, like the eigenspaces of the corresponding shape laplacians.
Moreover, when treated as a tool implementing intrinsic smoothing \cite{bensaid2023partialpiecewise} rather than a functional map network layer \cite{litany2017deep}, one could harness the power of the Laplace-Beltrami operator (LBO) spectral representation for state-of-the-art solutions for shape matching.

The proposed network optimization process involves a two term loss. 
The core, inspired by \cite{halimi2019unsupervised, bronstein2006generalized,memoli2005theoretical,bronstein2009partial,bronstein2006efficient},
utilizes the near isometry property of surfaces in nature as a measure of correspondence that can serve as an unsupervised/semi-supervised way of training our network. 
This loss measures the distortions of geodesic distance between corresponding points, when mapping one shape to the other. 
For the second regularization-term we suggest two penalizing options, related to the area preserving mapping \cite{roufosse2019unsupervised}, and an option to avoid the need to compute the functional map altogether.

We test the proposed framework on the SHREC'16 dataset \cite{cosmo2016shrec}, containing two different types of shape partiality benchmarks, cuts and holes.
We also evaluate it on our new dataset, PFAUST, with two different densities of holes.
We show that the proposed method has the lowest error in comparison to previous unsupervised partial shape matching methods.

\noindent\textbf{Contributions.}
\begin{enumerate}
    \item We prove that for partial shape matching a {\em functional map layer} introduces unavoidable errors into the resulting correspondence function. 
    \item We introduce PFAUST, a new benchmark that includes two datasets with different densities of holes, dedicated to partial-to-full shape correspondence.
    \item  We present a novel approach for partial shape matching avoiding the {\em functional map layer} altogether. 
    The proposed method reaches state-of-the-art results on the SHREC'16 and PFAUST by a large margin.
\end{enumerate}
\begin{figure}
    \centering
 \includegraphics[trim={0 0 0 2.1cm}, clip, width=\textwidth]{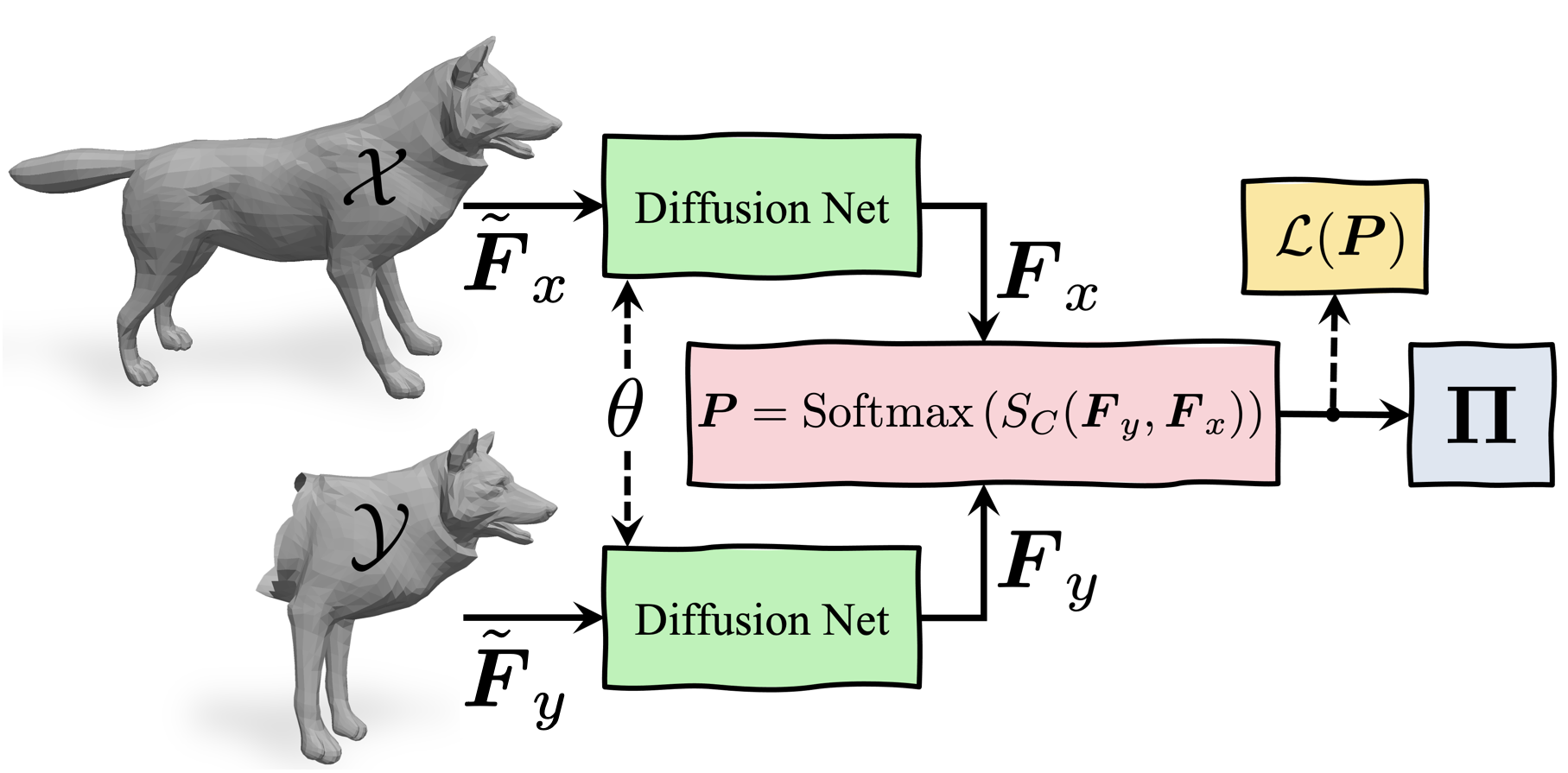}
    \caption{Overview of the proposed pipeline. 
    Basic features computed for the full and partial shapes are refined using a Siamese neural diffusion feature extractor. 
    Next, cosine similarity is computed and fed into a softmax layer that produces a soft correspondence. 
    While training, the loss functions are applied to the soft correspondence. 
    At inference, the soft correspondence matrix is binarized for sharp matching.}
    \label{fig:pipline}
\end{figure}
\section{Related efforts}
\label{related_work}
Deep learning was first introduced for dense shape correspondence in \cite{litany2017deep} proposing a supervised approach known as FM-Net, based on a differentiable functional map layer. 
Unsupervised versions of the FM-Net appeared in \cite{halimi2019unsupervised,roufosse2019unsupervised}.
To transform the supervised approach \cite{litany2017deep} into an unsupervised one, \cite{halimi2019unsupervised} suggested a loss based on the Gromov-Wasserstein Distance. 
That distance was suggested as a measure of choice for defining the correspondence of surfaces in nature, see \cite{elad2003bending,memoli2012some,bronstein2006generalized,bracha2020shape,pazi2020unsupervised,ginzburg2020cyclic, eisenberger2021neuromorph}. 
In another example, it was used to align the eigenfunctions of the scale invariant LBO of two shapes, see \cite{bracha2020shape}.
Here, we use the Gromov-Wasserstein loss without resorting to the functional map layer.

Later papers suggested new architectures \cite{sharp2022diffusionnet, geomfmnet,sharma2020weakly,li2020shape} improving the accuracy on networks trained in both supervised and unsupervised settings. 
In \cite{marin2020correspondence}, it was argued that learning robust basis functions instead of the LBO eigenfunctions improves correspondence results.
In \cite{srfeat}, it was suggested to avoid the functional map framework for correspondence of full shapes, relying instead on a supervised contrastive loss. 
The correspondence is done directly between the output network features, while a Dirichlet energy is minimized as part of the optimization to enforce smoothness provided by functional map methods.
Recently, a method for shape correspondence \cite{cao2023unsupervised} incorporates a loss \cite{attaiki2023understanding}, penalising the difference between the estimated functional map from the functional map layer, and the functional map extracted from soft correspondences calculated by cosine similarity between the output features of the network.
It was noticed in \cite{DPFM} that when matching partial shapes the functional map framework requires some assistance. 
Motivated by this observation, a network was introduced with an added attention component that makes the feature extractor aware of features on the other shape.
The network outputs vanilla features like other methods, those are then fed to the attention layer that refines the features.
Theoretically, features of points on the full shape that do not belong to the matched part can be zeroed out with this approach.
And yet, in our experiments we noticed that this method is hard to train. 
Indeed, in \cite{DPFM} only the non-refined features, without the attention part, were used for the functional map on the {SHREC'16-CUTS} benchmark.
In \cite{pfm}, relations between eigenvalues, eigenfunctions, and the functional map matrix for partial shapes are studied. 
Unlike \cite{pfm}, we explore the error introduced by shape partiality to the functional map matrix produced by the FM-layer, which is a least squares estimate of the matrix extracted from the ground truth correspondence.
While the error in \cite{pfm} relates to the length of the cut, our analysis exhibits a relation to the missing area.

\section{ Functional maps and shape partiality}

In 
\cref{FM_section}
we briefly provide the background for \textit{functional maps}, and in \cref{sec: shape partiality} and in the supplementary we analyse its issues with partial shape matching.

\subsection{Functional maps} \label{FM_section}
The notion of {\it functional maps} was introduced to non-rigid shape correspondence in \cite{fm}.
Given $T: \mathcal{X}\to \mathcal{Y}$,  a bijective mapping between two surfaces $\mathcal{X}$ and $\mathcal{Y}$, then $T$ induces a natural functional mapping $T_\mathcal{F}:\mathcal{F}(\mathcal{X}, \mathbb{R}) \to \mathcal{F}(\mathcal{Y}, \mathbb{R}), f\mapsto g = f\circ T^{-1}$ between scalar functions defined on each surface. 
Given two orthonormal bases $\{\psi_i^\mathcal{X}\}$ and $\{\psi_i^\mathcal{Y}\}$ of $\mathcal{F}(\mathcal{X},\mathbb{R})$ and $\mathcal{F}(\mathcal{Y},\mathbb{R})$,
we can decompose any function $f = \sum_i a_i\psi_i^\mathcal{X}\in\mathcal{F}(\mathcal{X},\mathbb{R})$ and its corresponding function $g = T_\mathcal{F}(f) = \sum_j b_j\psi_j^\mathcal{Y}$ in these bases.
Remarkably, the coefficients $(a_i)$ and $(b_j)$ are related by a linear transformation depending only on the mapping $T$ and the choice of bases,
%\begin{eqnarray}
 $   b_j = \sum\limits_{i} C_{ji} a_i$,
%\end{eqnarray}
where the coefficients $C_{ij} = \langle \psi_i^{\mathcal{Y}}, T_\mathcal{F}(\psi_j^\mathcal{X}) \rangle_\mathcal{Y}$ define the  {\it functional map}.

When $T$ is unknown, finding ${\boldsymbol C}=\{C_{ij}\}$ usually requires solving a least squares optimization problem, with constraints imposed by corresponding descriptor functions 
$(f_{i}^{\mathcal{X}},f_{i}^{\mathcal{Y}})$
defined on the surfaces $\mathcal{X}$ and $\mathcal{Y}$.
Defining 
${\hat{\bf F}}_x=\langle \psi_{i}^{\mathcal{X}},f_{j}^{\mathcal{X}}\rangle_{\cal{X}}$
and 
${{\hat{\bf F}}_y=\langle\psi_{i}^{\mathcal{Y}},f_{j}^{\mathcal{Y}}\rangle_{\cal{Y}}}$,
the matrix $\boldsymbol{C}$ can be computed by solving
\begin{eqnarray}
\label{CF=H}
\boldsymbol C &=&\argmin_{\boldsymbol{C}} \norm{ \boldsymbol C \hat{\boldsymbol F}_x - {\hat{\boldsymbol F}_y}}_F.
\end{eqnarray}
There is no restriction of which basis to choose.
However, it was shown in \cite{aflalo2015optimality} that selecting the leading eigenfunctions of the LBO, when ordered by their corresponding eigenvalues small to large, optimally and uniquely approximate the family of smooth functions with bounded Dirichlet energy.
When mentioning basis functions, eigenfunctions, or eigenvectors in this paper we refer to these low pass subsets of LBO leading basis elements. 

The least squares problem Eq.~(\ref{CF=H}) 
is solved by pseudo-inversion.
In deep learning frameworks, this operation is called the functional map layer (FM-layer) \cite{litany2017deep}, and
 deep learning pipelines for shape correspondence often incorporate it
\cite{roufosse2019unsupervised,DPFM,cao2023unsupervised,srfeat,halimi2019unsupervised}.
The FM-layer computes the solution to Eq.~(\ref{CF=H}) as
\begin{eqnarray}
    \label{eq: estimated FM pseudo inverse}
    \boldsymbol C_{yx} &= & \boldsymbol{\hat F}_x\boldsymbol{\hat F}_y^\top(\boldsymbol{\hat F}_y\boldsymbol{\hat F}_y^\top)^{-1}.
\end{eqnarray}
Here, $\boldsymbol C_{yx}$ is a spectral mapping between $\mathcal{Y}$ and $\mathcal{X}$.
In the discrete setting,
denoting $\boldsymbol F_x$ and $\boldsymbol F_y$ the corresponding feature matrices on $\mathcal{X}$ and $\mathcal{Y}$,
this becomes 
{%\small
\begin{eqnarray}
    \boldsymbol C_{yx} & = & \boldsymbol \Psi_x^\top \boldsymbol A_x \boldsymbol F_x \boldsymbol F_y^\top\boldsymbol A_y\boldsymbol \Psi_y(\boldsymbol \Psi_y^\top\boldsymbol A_y\boldsymbol F_y\boldsymbol F_y^\top\boldsymbol A_y\boldsymbol \Psi_y)^{-1},
    \label{discrite_fm}
\end{eqnarray}
}%
where $\boldsymbol{\Psi}_x$ 
is the matrix of basis functions on $\mathcal{X}$ and
$\boldsymbol A_x$ is
the diagonal matrix of the areas about 
%the
vertices
of the triangulated surface $\mathcal{X}$.
For conciseness,
we 
write
\begin{eqnarray}
    \boldsymbol{Q}_y &=& \boldsymbol \Psi_y^\top\boldsymbol A_y\boldsymbol F_y\boldsymbol F_y^\top\boldsymbol A_y\boldsymbol \Psi_y.
\end{eqnarray} 

%____________________________________
\subsection{Shape partiality}
\label{sec: shape partiality}
Most deep learning approaches for shape correspondence use the same three  components.
First, a neural network generates features at each point of the surface. 
Then, these features are fed to an FM-layer to compute the functional map between the surfaces.
Note, that this is the first time that the features of the two shapes interact.
Finally, the functional map is used to estimate a pointwise correspondence between the two shapes. 
We prove in this subsection that in the case of partial shapes, the use of an FM-layer introduces unavoidable errors.

We begin our analysis with the continuous case.
%________________________________
\begin{theorem}[Continuous case]
\label{th: FM error continuous case}
Given two surfaces $\mathcal{X}$ and $\mathcal{Y}$, where $\mathcal{Y}\subset\mathcal{X}$.
Denote $\mathcal{Z} = \mathcal{X}\setminus\mathcal{Y} $, and
let $\{\psi_i^{\cal X}\}_{i=1}^k$ and $\{\psi_i^{\cal Y}\}_{i=1}^k$ be subsets of orthonormal bases of $\mathcal{X}$ and $\mathcal{Y}$, respectively. 
Let $\hat{\boldsymbol{F}}_x$ and $\hat{\boldsymbol{F}}_y$ be the coefficients of the corresponding feature functions on the corresponding surfaces. 
Then, the functional map layer between $\mathcal{X}$ and $\mathcal{Y}$ computes a functional map $ \hat{\boldsymbol{C}}_{yx} = \boldsymbol C_{yx} + \boldsymbol C_{yx}^E$,  where
\begin{eqnarray}
    \boldsymbol  C_{yx}=  \hat{\boldsymbol {F}}^{(\mathcal{Y})} \hat{\boldsymbol F}_y^\top(\hat{\boldsymbol F}_y\hat{\boldsymbol F}_y^\top)^{-1}
    \quad\quad\text{and}\quad\quad
    \boldsymbol  C_{yx}^E &=&  \hat{\boldsymbol {F}}^{(\mathcal{Z})} \hat{\boldsymbol F}_y^\top(\hat{\boldsymbol F}_y\hat{\boldsymbol F}_y^\top)^{-1}.
\end{eqnarray}
    Here, the superscripts ${(\mathcal{Y})}$ and ${(\mathcal{Z})}$ indicate a matrix containing only information related to regions of $\mathcal{X}$ that correspond to the partial surfaces $\mathcal{Y}$ or $\mathcal{Z}$, respectively.
\end{theorem}
%____________________________
%____________________________
A proof can be found in the supplementary.
The matrix $\boldsymbol{C}_{yx}$ is the  functional map relating $\boldsymbol{\Psi}^\mathcal{X}$ and $\boldsymbol{\Psi}^\mathcal{Y}$ when considering only descriptors which vanish on $\mathcal{Z}$. 
We show in the supplementary that it is the functional map  deduced from a perfect mapping between $\mathcal{Y}$ to $\mathcal{X}$.
At the other end,  $ \boldsymbol C_{yx}^E$ is an error term that can be interpreted as the  functional map relating $\boldsymbol{\Psi}^\mathcal{X}$ and $\boldsymbol{\Psi}^\mathcal{Y}$ when considering only descriptors that vanish on $\mathcal{X}\setminus \mathcal{Z}$.
We readily deduce the following corollary,
%__________________________
\begin{corollary}
If $\hat{\boldsymbol {F}}^{(\mathcal{Z})} \hat{\boldsymbol F}_y^\top \neq \boldsymbol{0}$, then, the functional map layer injects an error $\boldsymbol C_{yx}^E\neq \boldsymbol 0$ into the functional map.
\end{corollary}

\begin{figure}[ht]
\centering 
{
\resizebox{\textwidth}{!}{
        \includegraphics[width=0.324\linewidth]{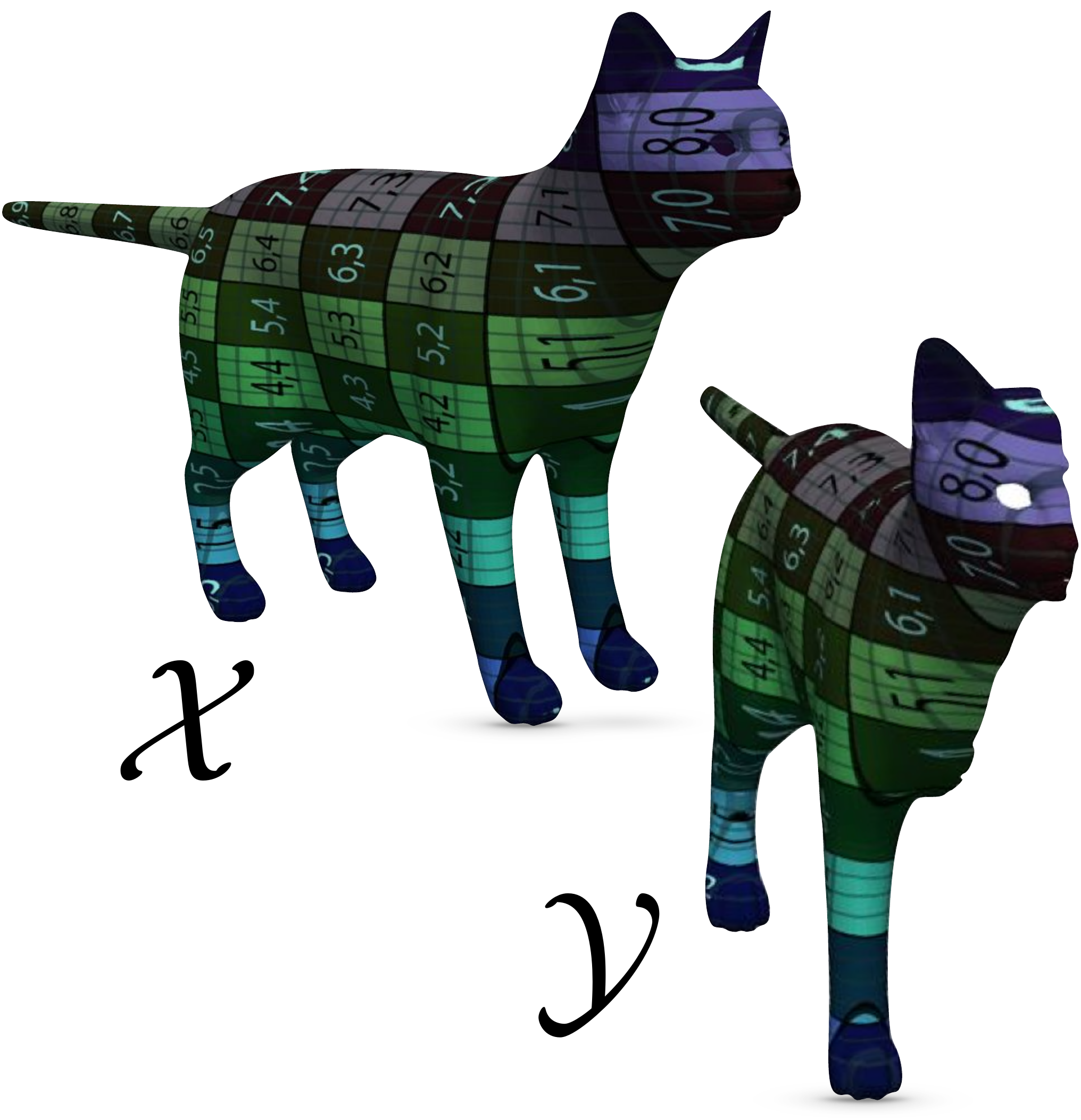}
        \quad
        \includegraphics[width=0.313\linewidth]{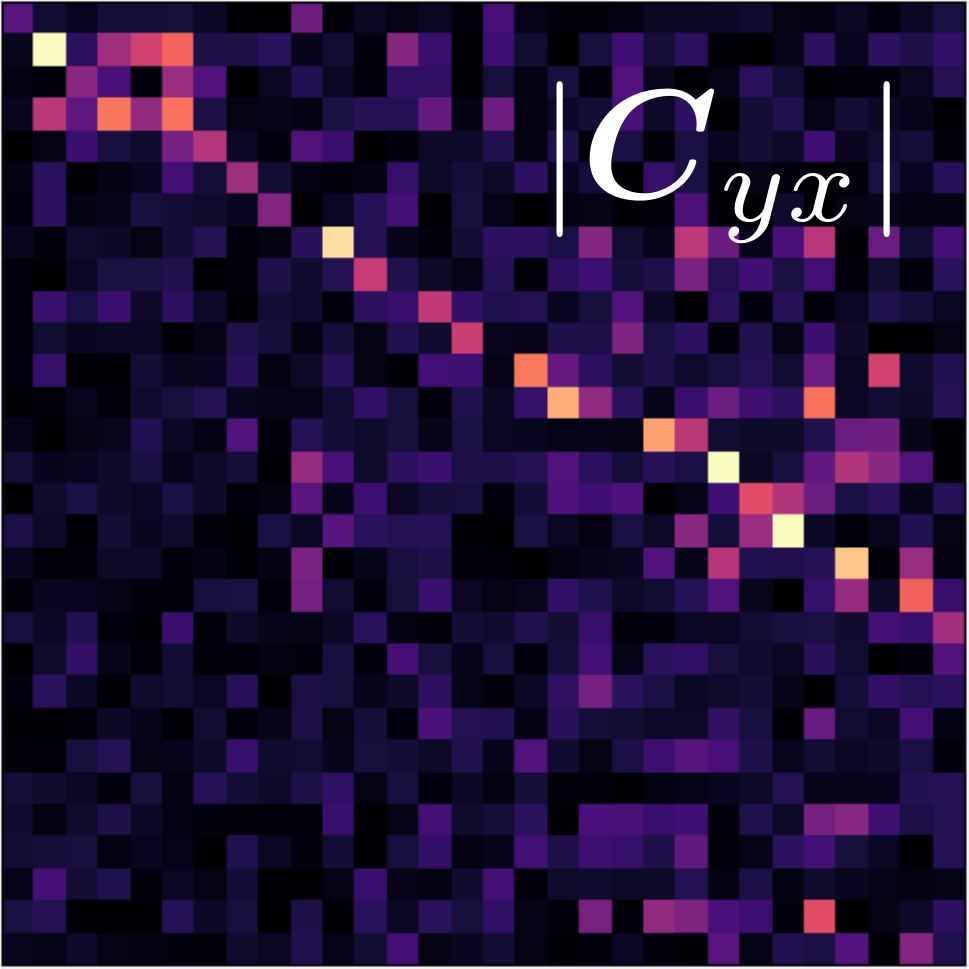}  
        \hspace{0.25em}
        \includegraphics[width=0.391\linewidth]{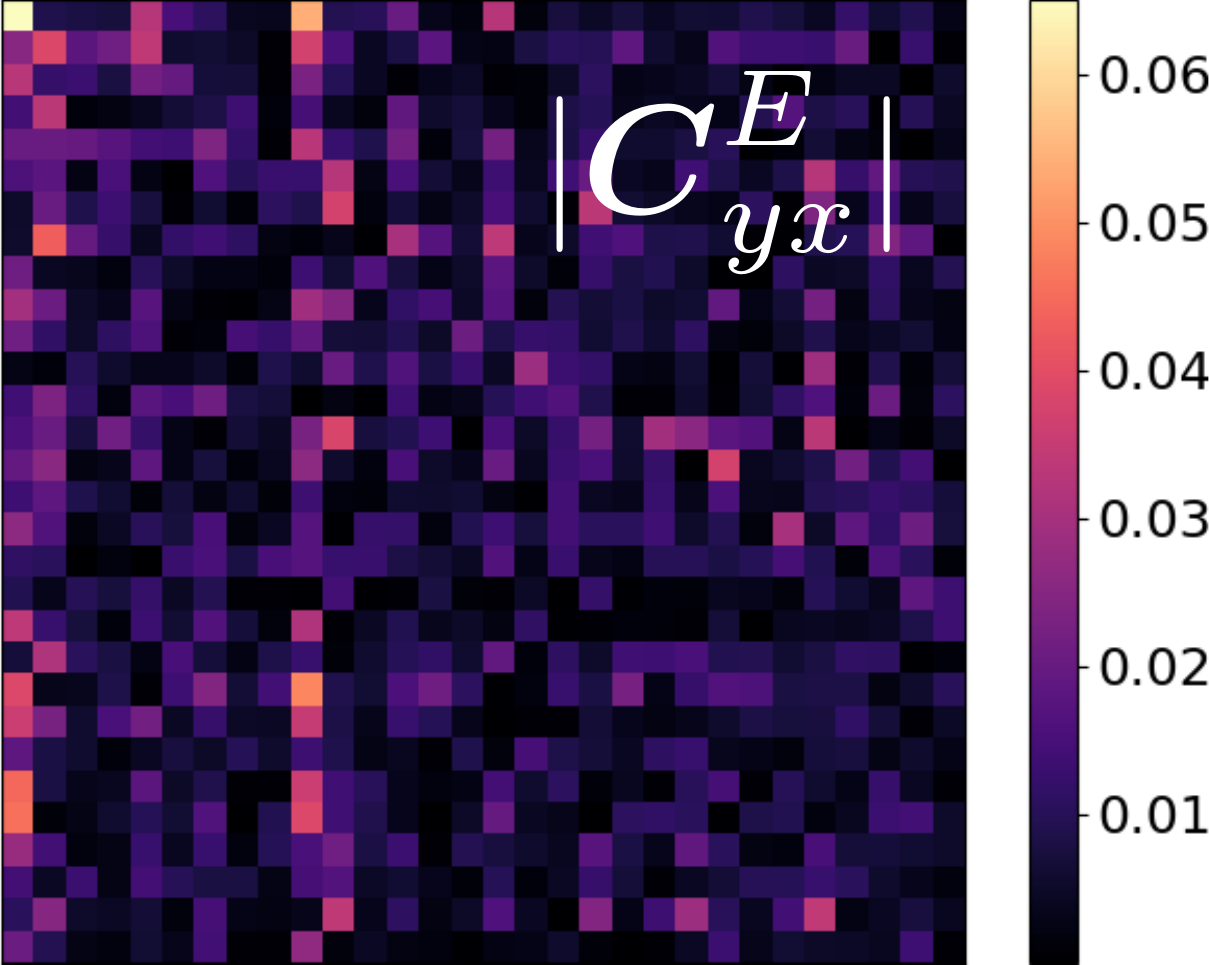}
 }
}
 \caption{
    Estimating the functional map $\hat{\boldsymbol{C}}_{yx}$ between a full $\mathcal{X}$ and its part $\mathcal{Y}$  using features independently extracted for each yields errors.
    Recall, $\hat{\boldsymbol{C}}_{yx} = \boldsymbol{C}_{yx} + \boldsymbol{C}_{yx}^E$, where $\boldsymbol{C}_{yx}$ is the ideal functional map given the correct matching, and $\boldsymbol{C}_{yx}^E$ is an error resulting from matching the part $\mathcal{Y}$ to its complementary part in $\mathcal{X}$.
    Here, we plot the magnitude of entries in $|\boldsymbol{C}_{yx}|$ and $|\boldsymbol{C}_{yx}^E|$. 
    Indeed, $\boldsymbol{C}_{yx}$ contains an informative structure, whereas $\boldsymbol{C}_{yx}^E$ is a noise-like structure-less matrix.
    }
  \label{fig:functional map matrix decomposition}
\end{figure}

Generally, $\hat{\boldsymbol {F}}^{(\mathcal{Z})} \hat{\boldsymbol F}_y^\top \neq \boldsymbol{0}$, thus, the functional map layer unavoidably introduces a significant error to the estimated functional map.
The above theorem also applies to the discrete case, a proof can be found in the supplementary..
\begin{theorem}[Discrete case]
\label{th: discrete case}
Given a sampled surface $\mathcal{X}$, like a triangle mesh, split into two subsurfaces $\mathcal{Y}$ and $\mathcal{Z}$.
The functional map layer between $\mathcal{X}$ and $\mathcal{Y}$ yields a functional map $ \hat{\boldsymbol{C}}_{yx} = \boldsymbol C_{yx} + \boldsymbol C_{yx}^E$, where
\begin{eqnarray}
    \label{c_yx}
    {\small
    \boldsymbol  C_{yx}=  (\boldsymbol \Psi^{(\mathcal{Y})})^\top\boldsymbol A_y\boldsymbol F^{(\mathcal{Y})}\boldsymbol F_y^\top \boldsymbol A_y\boldsymbol \Psi_y\boldsymbol{Q}^{-1}_y
    \;\mbox{and}\;
    \boldsymbol C^E_{yx} = (\boldsymbol \Psi^{(\mathcal{Z})})^\top\boldsymbol A_z\boldsymbol F^{(\mathcal{Z})}\boldsymbol F_y^\top \boldsymbol A_y\boldsymbol \Psi_y\boldsymbol{Q}^{-1}_y.}
\end{eqnarray}%

\end{theorem}%
% A proof can be found in the supplementary.
%
\begin{corollary}
If the feature dimensions
${d < \min\{n_y,n_z\}}$, where $n_y$ and $n_z$ are the number of sampled vertices of surfaces $\mathcal{Y}$ and $\mathcal{Z}$,
%respectively, 
and if both $\boldsymbol F^{(\mathcal{Z})}$ and $\boldsymbol F_y$ have $d$ rank, and if both\footnote{Where $\perp_{M}$  means orthogonality with respect to the metric matrix $M$.} 
$\boldsymbol F_y \not\perp_{A_y}  \boldsymbol{\Psi}_y$ and
${\boldsymbol F^{(\mathcal{Z})} \not\perp_{A_z}  \boldsymbol{\Psi}^{(\mathcal{Z})}}$, then, $\boldsymbol C^E_{yx} \neq \boldsymbol{0}$. 
\end{corollary}
In practice, $d \ll \min\{n_y,n_z\}$ and the other assumptions are satisfied. 
This means that
for
partial shape matching, the FM-layer unavoidably introduces an error into the estimated functional map.
Nevertheless, as claimed in \cite{DPFM}, it is possible for the error to vanish. 
Unfortunately, this can only happen when $d = n_x$, that is, in the unreasonable case where the descriptor dimensionality is equal to the total number of sampled vertices.
Obviously, taking $\boldsymbol F^{(\mathcal{Z})} \rightarrow \boldsymbol 0$ or 
${\boldsymbol F^{(\mathcal{Z})} \perp_{A_z}  \boldsymbol{\Psi}^{(\mathcal{Z})}}$
would eliminate this error; however, any general feature extractor would not be aware of the given shape partiality, that is, the matching region of $\mathcal{Y}$  in $\mathcal{X}$.   
As mentioned, the features are extracted independently for each shape without any interaction before the FM-layer. 
As such, we cannot  have a general meaningful feature extractor that provides $\boldsymbol 0$ on $\mathcal{Z}$. 
See \cref{fig:functional map matrix decomposition} for an example of the computed matrices $\boldsymbol{C}_{yx}$ and $\boldsymbol{C}_{yx}^E$.
Methods such as \cite{pfm} and those relying on \cite{fm_mask} deal with shape partiality by masking the matrix $\boldsymbol{C}$.
One could easily generalise the proposed analysis along this line of thought, by which one introduces an additive mask matrix to $\boldsymbol{Q}$.
We explore in detail the error induced by the FM-layer in the supplementary. 
We show that under simplifying assumptions  the error of the FM-layer is proportional to the missing area.

%_________________________________
\section{Method}
\label{sec: method}
\begin{figure}[tbp]
  \centering
   \includegraphics[width=\textwidth]
   {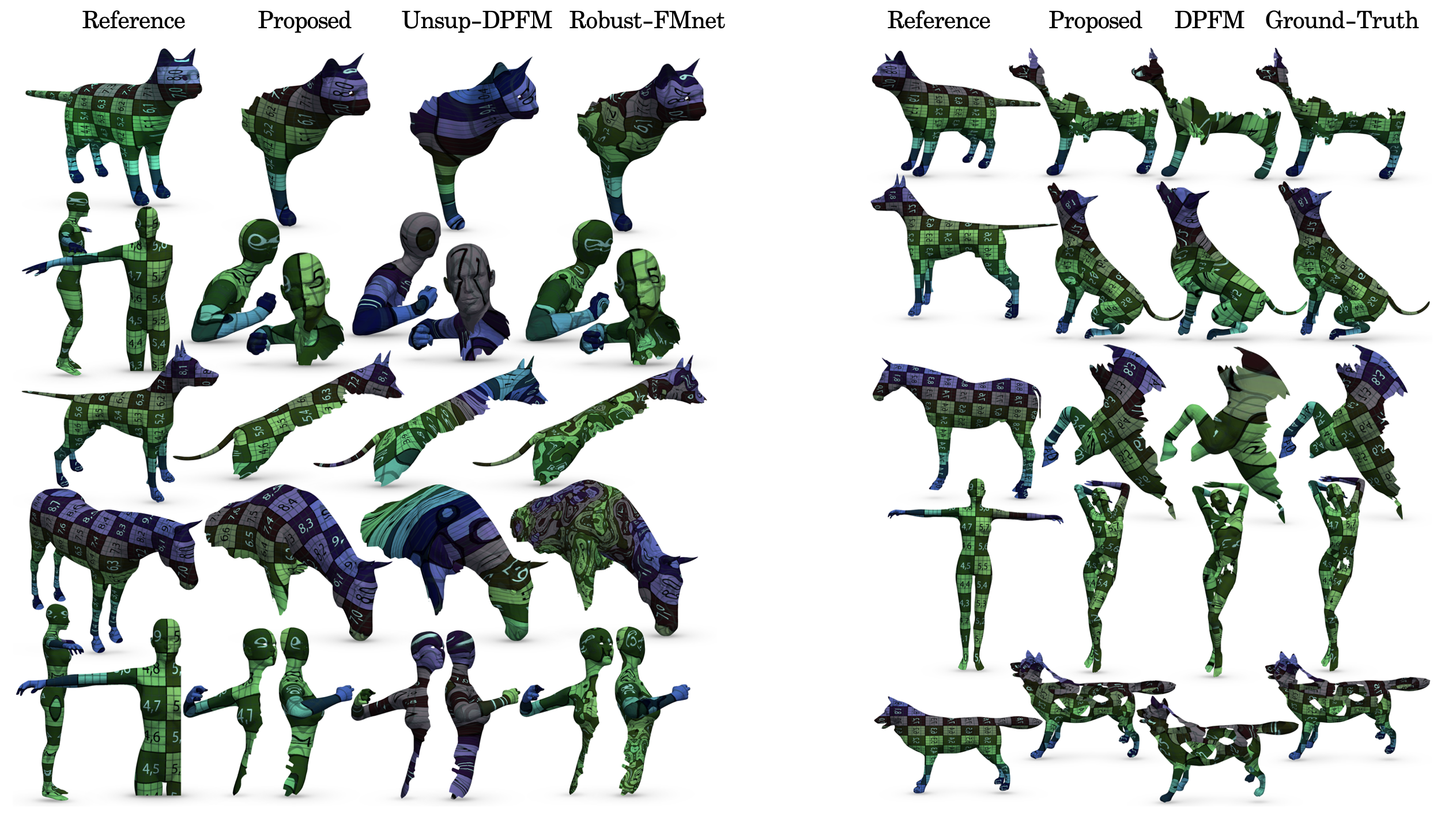}
   \caption{
   Qualitative results on SHREC'16 CUTS (\emph{left}) and HOLES (\emph{right}).
   On SHREC'16 CUTS, we obtain visually appealing results that outperform previous unsupervised methods. 
   On SHREC'16 HOLES, we obtain better matching results both visually and quantitatively than even the best supervised approach (DPFM). %Please zoom in.
   }
   \label{fig:qualitative_result_12}
\end{figure}
We have proven that the FM-layer involves an unavoidable error when applied to partial shape matching. 
Utilizing this knowledge, we propose to avoid an FM-layer and apply a simple and effective approach, see Figure \ref{fig:pipline}. 

Differentiable softmax of point features' inner product in the context of shape matching were first introduced in
\cite{cao2023unsupervised,attaiki2023understanding}.
We follow a similar line of thought, and estimate a soft correspondence matrix directly from the feature similarity of the two shapes using the softmax operator. 
Unlike the methods suggested in \cite{cao2023unsupervised,attaiki2023understanding}, we do not feed the features from the feature extractor to an FM-layer.
The correspondence is extracted using only feature similarity.

This correspondence matrix is an input to our loss functions. 
The main component of the proposed loss was first used in \cite{halimi2019unsupervised} in the context of learning surface matching.
It penalizes the distortion of surface inter-geodesic distances between corresponding pairs of points.
For regularization, we propose two options.
The first option \cite{roufosse2019unsupervised} is an orthogonality regularization of the estimated functional map extracted from the soft correspondence matrix,
but it does not involve an FM-layer.
It allows to apply a low pass filter on the computed correspondence in the dual spectral domain. 
To compute the functional map, we simply multiply the correspondence matrix by the leading LBO basis functions.
We also propose an alternative regularization option. 
It is based on the assumption that the low frequency or leading eigenfunctions on $\mathcal{X}$ restricted to the regions corresponding to $\mathcal{Y}$ could be well approximated by the low frequency leading eigenfunctions on $\mathcal{Y}$.
This 
option avoids the need to compute the functional map altogether.

%______________________________________
\paragraph{Feature extractor.}
A feature extractor in our context is a neural network that considers as
input a given feature vector $\Tilde{\boldsymbol{F}} \in \mathbb{R}^{n\times \Tilde{d} }$ and outputs refined features $\boldsymbol{F}\in \mathbb{R}^{n\times d}$, where $n$ is the number of vertices of the triangle surface, and $\Tilde d$ (resp. $d$) is the number of given (resp. refined) descriptor elements. 
An example of input features are the coordinates in $\mathbb{R}^3$ of each vertex of a given surface. 
Following a Siamese philosophy, the same network, is applied to both shapes.

%________________________________________________
\paragraph{The correspondence matrix $\boldsymbol{P}$.}
Given the extracted features for each shape, we compute their pointwise cosine similarity 
\begin{eqnarray}
S_C(\boldsymbol F_y, \boldsymbol F_x) &\equiv & \left \langle \frac{\boldsymbol F_y}{\lVert\boldsymbol F_y\rVert}, \frac{\boldsymbol F_x}{\lVert\boldsymbol F_x\rVert} \right \rangle.
\end{eqnarray}
We then apply a softmax operator to estimate directly a soft-corresponding matrix $\boldsymbol{P}_{yx}$ mapping shape $\mathcal{Y}$ to shape $\mathcal{X}$. 
Note, that this approach  differs from the common practice as we do not 
use an FM-layer to estimate correspondences.
$\boldsymbol{P}_{yx}$ is calculated as in \cite{cao2023unsupervised, bensaid2023partialpiecewise},
\begin{eqnarray}
    \boldsymbol{P}_{yx} &=& \text{Softmax}\left(\tau^{-1} S_C(\boldsymbol F_y, \boldsymbol F_x)\right),
\end{eqnarray}
where $\boldsymbol F_x$ and $\boldsymbol F_y$ are the refined features of the full shape $\mathcal{X}$ and partial shape $\mathcal{Y}$ respectively, $\tau$ is a temperature hyper-parameter, 
and
\begin{eqnarray}
    \text{Softmax}(\boldsymbol{X})_{ij} &=& \frac{\text{exp}(\boldsymbol{X}_{ij})}{\sum_j \text{exp}(\boldsymbol X_{ij})}.
\end{eqnarray}

Note, that by choosing the softmax operator we only derive the correspondence from the partial shape $\mathcal{Y}$ to its full version $\mathcal{X}$, as some regions in $\mathcal{X}$ do not have matching regions in $\mathcal{Y}$.
Obviously, the partiality requirement hinders the use of bidirectional loss functions.
As elaborated in the next subsection, some published losses require both $\boldsymbol{P}_{xy}$ and $\boldsymbol{P}_{yx}$, see for example, \cite{roufosse2019unsupervised}.
We discuss further advantages of the softmax operator compared to the FM-layer in \cref{sec:The Softmax operator compared to the FM-Layer}.

During training, we use regularization based on the functional map computed from the correspondence matrix by
\begin{eqnarray}
    \boldsymbol{C}_{yx} &=& \boldsymbol\Psi_y^\top\boldsymbol A_y \boldsymbol{P}_{yx} \boldsymbol{\Psi}_x.
    \label{p2c}
\end{eqnarray}
As in \cite{cao2023unsupervised}, at inference, we only compute the point-to-point matching $\boldsymbol{\hat\Pi}_{yx}^{\text{iso}}$, by
\begin{eqnarray}
    \boldsymbol{\hat\Pi}_{yx}^{\text{iso}}(i) & = & \argmax_j 
\,
(\boldsymbol{\Psi}_y\boldsymbol{\Psi}_y^\top\boldsymbol{A}_y \boldsymbol{\hat\Pi}_{yx}\boldsymbol{A}_x\boldsymbol{\Psi}_x\boldsymbol{\Psi}_x^\top)_{ij},
\end{eqnarray}%
where $\boldsymbol{\hat\Pi}_{yx} = \text{NN}(\boldsymbol F_y,\boldsymbol F_x)$, and NN is the nearest neighbor operator.

%________________________________________

\paragraph{Unsupervised learning to match.}
The proposed pipeline operates mainly in the spatial domain rather than the spectral one. 
The core loss, generalising \cite{halimi2019unsupervised} to partial shapes, is based on the Gromov distance \cite{memoli2005distance,bronstein2010gromov,bronstein2006generalized, sgmds}.
Our loss measures the quality of the partial shape correspondence matrix $\boldsymbol{P}_{yx}$. 
It is given by,
\begin{eqnarray}
\label{eq: L dist}
    \mathcal{L}_\text{G} &=& \norm{\boldsymbol{A}^{\frac{1}{2}}_y\big(\boldsymbol{P}_{yx}^\top \boldsymbol D_x \boldsymbol{P}_{yx} - \boldsymbol{D}_y\big)\boldsymbol{A}^{\frac{1}{2}}_y}^2_F,
\end{eqnarray}
where $\boldsymbol{D}_x$ and $\boldsymbol{D}_y$ are the inter-geodesic distance matrices of $\mathcal{X}$ and $\mathcal{Y}$.
The $(i,j)$ entry of $\boldsymbol{D}_x$ is the geodesic distance on $\mathcal X$ between its $i$-th and $j$-th vertices. 

This loss promotes the preservation of inter-point geodesic distances, which should hold if the surfaces are related by an isometric deformation.
This property applies for most articulated objects in nature. 
% ---------------------------------

Next, we study the unsupervised regularization loss
\begin{eqnarray}
\label{eq: L orth}
    \mathcal{L}_\text{orth} &=& \norm{\hat{\boldsymbol{C}}_{yx} \hat{\boldsymbol{C}}_{yx}^\top - \boldsymbol{J}_r}^2_F.
\end{eqnarray}
Here, $r$ is the number of chosen eigenfunctions of the LBO of $\mathcal{Y}$ that correspond to smaller than the largest eigenvalue of the chosen eigenbasis of the LBO of $\mathcal{X}$.
Note, by selecting a small finite number of eigenfunctions, we actually keep the low-frequencies.
This measure promotes the orthogonality of functional maps, which should
hold when the mapping between shapes is area preserving \cite{fm, DPFM}.

Plugging Eq.~(\ref{p2c}) into Eq.~(\ref{eq: L orth}) yields
\begin{eqnarray}
     \mathcal{L}_\text{orth} 
    & = & \norm{\boldsymbol\Psi_y^\top\boldsymbol A_y \boldsymbol{P}_{yx} \boldsymbol{\Psi}_x \boldsymbol{\Psi}_x^\top  \boldsymbol{P}_{yx}^\top \boldsymbol A_y \boldsymbol\Psi_y - \boldsymbol{J}_r}^2_F.
\end{eqnarray}
Note, that computing $\mathcal{L}_{\text{orth}}$ requires explicit estimation of the functional map from the soft correspondence. 
Let, $\tilde{\mathcal{L}}_\text{orth}$ be the following modified regularization,
\begin{eqnarray}
    \tilde{\mathcal{L}}_\text{orth} &=& 
     \norm{\boldsymbol{\Psi}_y(\boldsymbol C_{yx} \boldsymbol{C}_{yx}^\top - \boldsymbol{J}_r)\boldsymbol{\Psi}_y^\top}^2_F \cr
          &=&
    \norm{\boldsymbol H_y\boldsymbol{P}_{yx} \boldsymbol{\Psi}_x \boldsymbol{\Psi}_x^\top  \boldsymbol{P}_{yx}^\top \boldsymbol H_y^\top - \boldsymbol\Psi_y\boldsymbol{J}_r\boldsymbol\Psi_y^\top}^2_F
    \cr
     &=& \norm{\boldsymbol H_y\boldsymbol{P}_{yx} \boldsymbol{\Psi}_x \boldsymbol{\Psi}_x^\top  \boldsymbol{P}_{yx}^\top \boldsymbol H_y^\top - \boldsymbol{\Tilde{\Psi}}_y\boldsymbol{\Tilde{\Psi}}_y^\top}^2_F,
\end{eqnarray}
where $\boldsymbol{H}_y = \boldsymbol{\Psi}_y\boldsymbol{\Psi}_y^\top\boldsymbol A_y$ and  $\boldsymbol{\Tilde{\Psi}}_y = \boldsymbol\Psi_y\boldsymbol{J}_r$. 
$\boldsymbol{H}_y$ is the projection operator onto the set of low frequency basis functions $\boldsymbol{\Psi}_y$.
Thus, $\boldsymbol{H}_y$ is a geometric low pass filter.

If $\boldsymbol{P}_{yx}$ is the true mapping matrix from $\mathcal{X}$ to $\mathcal{Y}$,
then the leading low frequency eigenvectors $\boldsymbol{\Psi}_x$, restricted to 
$\mathcal{Y}$,  would be well captured by the low frequencies corresponding to
leading LBO eigenvectors on $\mathcal{Y}$, 
\cite{aflalo2015optimality}. 
In that case, 
\begin{eqnarray}
    \label{eq:BPXtoY}
    \boldsymbol H_y\boldsymbol{P}_{yx} \boldsymbol{\Psi}_x &=& \boldsymbol H_y \boldsymbol{\Psi}_x^y \,\,\approx \,\, \boldsymbol{\Psi}_x^y,
\end{eqnarray}
where $\boldsymbol{\Psi}_x^y$ is the restriction of  $\boldsymbol{\Psi}_x$ to its part corresponding to $\mathcal{Y}$.
Considering these relations, we propose  the {\it low-pass filter} loss
\begin{eqnarray}
    \mathcal{L}_\text{lpf}   &=&
    \norm{\boldsymbol{P}_{yx} \boldsymbol{\Psi}_x \boldsymbol{\Psi}_x^\top  \boldsymbol{P}_{yx}^\top - \boldsymbol{\Tilde{\Psi}}_y\boldsymbol{\Tilde{\Psi}}_y^\top}^2_{\mathcal{Y},\mathcal{Y} }, 
     \label{eq: L lpf}
\end{eqnarray}
where $ \norm{\boldsymbol{B}}_{\mathcal{Y},\mathcal{Y} } = \text{trace}(\boldsymbol{A}_y\boldsymbol B^T\boldsymbol A_y \boldsymbol B)$ as defined in \cite{sgmds}. 
When the relation in Eq.~(\ref{eq:BPXtoY}) holds, we readily have $\mathcal{L}_\text{lpf} \approx  \mathcal{L}_\text{orth}$. 
Note that $\mathcal{L}_\text{lpf}$ does not require the computation of the functional map matrix.

Our loss functions are the combination of  $\mathcal{L}_\text{G}$ coupled with a low pass regularization. 
We propose to use either \begin{eqnarray}
    \label{eq: TOTAL loss orth}
    \mathcal{L} &=&  \mathcal{L}_\text{G}  + \lambda_\text{orth}\mathcal{L}_\text{orth},
\end{eqnarray} 
which requires the computation of the functional map based on the soft correspondence matrix, or 
\begin{eqnarray}
    \label{eq: TOTAL loss lpf}
    \mathcal{L} &=&  \mathcal{L}_\text{G}  + \lambda_\text{lpf}\mathcal{L}_\text{lpf},
\end{eqnarray}
which bypasses functional maps altogether.
The coefficients $\lambda_\text{orth}$ and $\lambda_\text{lpf}$ are the relative weights of the $\mathcal{L}_\text{orth}$ and 
$\mathcal{L}_\text{lpf}$ regularization parts of the loss, respectively.

%___________________________________________
\paragraph{Refinement method.}
We follow the refinement process of \cite{cao2023unsupervised} and use test-time adaptation refinement.
Unlike other post-processing methods, this refinement adjusts the weights of the network during test-time for each pair of shapes via a 
few back-propagation iterations.
Unlike \cite{cao2023unsupervised}, we do not change the loss functions or add new ones during the refinement process.

%__________________________________
\paragraph{Implementation considerations.}
For a feature extractor we chose the DiffusionNet
model \cite{sharp2022diffusionnet}.
The chosen input features $\Tilde{\boldsymbol{F}}$ of our model are the 3D \textit{xyz}-coordinates concatenated with the surface normals. 
The output dimension of the refined features is $d = 256$, and we use four DiffusionNet blocks, where the inner spectral resolution and number of hidden channels are $128$. 
We set $\tau$ to be $0.07$ and $0.01$ for the SHREC'16 HOLES and CUTS benchmarks respectively.
For all experiments we set $\lambda_\text{orth} = 0.001$ and $\lambda_{\text{lpf}} = 0.02$.
Each shape uses a truncated LBO basis of $k=200$ eigenfunctions. 
We train the weights of the models using the ADAM optimizer  \cite{kingma2014adam} with a learning rate of $10^{-3}$. 
Similar to \cite{cao2023unsupervised}, we set the number of refinement iterations to $15$.
The distance matrices are calculated using the Fast-Marching Method \cite{fmm}.
%_______________________
\section{Experiments}
Here, we compare our method to current partial shape matching {techniques.\footnote{Code and data can be found at \url{https://github.com/ABracha/DirectMatchNet}
and  \url{https://github.com/ABracha/PFAUST}.}}

%___________________________________________
\paragraph{Datasets and evaluation method.}
The reference evaluation benchmarks for partial shape matching are the SHREC'16 \text{CUTS} and \text{HOLES} datasets \cite{cosmo2016shrec}.
Each dataset contains $8$ different shapes under various poses, which are non-rigid semi-isometric deformations of a reference pose.
The shapes in the \text{CUTS} dataset are obtained by cutting full shapes into two parts by a plane and keeping only one part, whereas those in the \text{HOLES} dataset are formed by applying  holes and cuts. 
The CUTS (resp. HOLES) dataset contains $120$ (res. 80) shapes in the training-set, and both datasets have $200$ shapes for testing.
Empirically, HOLES is believed to be a more challenging benchmark. 
Indeed, its shapes are trickier to handle for partial shape matching;  they involve, on average, longer boundaries while the number of triangles and vertices of the partial shape is smaller. 
When the amount of holes increases, the feature extractor struggles to produce consistent features common to the partial and full shapes.
In addition, we further apply the proposed method to a new benchmark we created by introducing holes to shapes in the FAUST remeshed \cite{faustRemeshed} dataset, in a similar way to SHREC'16 HOLES that were generated by introducing holes to shapes from TOSCA \cite{TOSCA}. 
We produce two datasets which we made publicly available, the  PFAUST-M and  PFAUST-H standing for \textit{medium} and \textit{hard} difficulties, the latter having more holes than the former. 
Each dataset contains human shapes from 10 different people, each in 20 poses, 10 for partial shapes and 10 for full shapes. 
More details on {PFAUST} are provided in the supplementary.
We follow the Princeton protocol \cite{kim2011blended} for evaluation, using the mean geodesic error, and as in other methods \cite{DPFM, cao2023unsupervised}, we normalize the square root of the area of the shapes to one.

%_________________________
\paragraph{Baselines.}
Baseline methods we compare to are grouped into three categories. 
First, axiomatic methods including PFM \cite{pfm} and FSP \cite{fsp}. 
Second, supervised methods including DPFM \cite{DPFM} and GeomFMaps \cite{geomfmnet}.
Third, unsupervised methods including unsupervised DPFM \cite{DPFM}
 and RobustFmaps \cite{cao2023unsupervised}. 
 Supervised DPFM is the current state-of-the-art approach. 
 We re-implemented the loss function of the unsupervised DPFM due to source code unavailability. 
The results presented for RobustFmaps differ from those shown in the original paper, since we did not pre-train on the TOSCA \cite{TOSCA} or other external datasets, see supplementary for details.
For post-processing refinement, we used the Zoomout method \cite{zoomout} and the time adaptation refinement \cite{cao2023unsupervised}.

\begin{table}[ht]
\caption{Quantitative results on SHREC'16. 
The numbers correspond to the mean geodesic error (scaled by 100), and the result using post processing refinement. 
Best performance is marked in bold and second best is underlined.
The results demonstrate the superiority of the proposed method in the unsupervised arena. 
It outperformed both supervised and unsupervised approaches on the HOLES benchmark, which contains extremely challenging scenarios of shape parts.
}
\label{tab:result_table}
\centering
\resizebox{\textwidth}{!}{
    \begin{tabular}{@{}c|c|c|c|c@{}}
    \toprule
     Test-set & \multicolumn{2}{c|}{\text{CUTS}} & \multicolumn{2}{c}{\text{HOLES}} \\
     \toprule
    Training-set  & \text{CUTS} & \text{HOLES}  & \text{CUTS} & \text{HOLES} \\
    \midrule
    \multicolumn{5}{c}{Axiomatic methods}\\
    \midrule
    PFM \cite{pfm}$\rightarrow$Zoomout & \multicolumn{2}{c|}{9.7 $\rightarrow$ 9.0}  &  \multicolumn{2}{c}{23.2 $\rightarrow$ 22.4} \\
    FSP \cite{fsp} $\rightarrow$ Zoomout &  \multicolumn{2}{c|}{16.1 $\rightarrow$ 15.2} & \multicolumn{2}{c}{33.7 $\rightarrow$ 32.7} \\
    \midrule
    \multicolumn{5}{c}{Supervised methods}\\
    \midrule
    GeomFMaps \cite{geomfmnet} $\rightarrow$ Zoomout  &  12.8 $\rightarrow$ 10.4 & 19.8 $\rightarrow$ 16.7 &  20.6 $\rightarrow$ 17.4 & 15.3 $\rightarrow$ 13.0 \\   
    DPFM \cite{DPFM} $\rightarrow$ Zoomout &  \textbf{3.2 $\rightarrow$ 1.8} & \textbf{8.6} $\rightarrow$ \underline{7.4} &  \underline{15.8} $\rightarrow$ 13.9 & 13.1 $\rightarrow$ 11.9\\   
    \midrule
    \multicolumn{5}{c}{Unsupervised methods}\\
    \midrule
    Unsupervised-DPFM \cite{DPFM} $\rightarrow$ Zoomout  &  11.8 $\rightarrow$ 12.8 & 19.5 $\rightarrow$ 18.7 &  19.1 $\rightarrow$ 18.3 & 17.5 $\rightarrow$ 16.2 \\
    RobustFMnet \cite{cao2023unsupervised} $\rightarrow$ Refinement &  16.9 $\rightarrow$10.6& 22.7$\rightarrow$16.6 &  18.7 $\rightarrow$ 16.2 & 23.5 $\rightarrow$ 18.8\\
     Proposed Orthogonal $\rightarrow$ Refinement  &  \underline{6.9} $\rightarrow$ 5.6 & 12.2 $\rightarrow$ 8.0  &  \textbf{14.2} $\rightarrow$ \textbf{10.2} & \textbf{11.4 $\rightarrow$ 7.9} \\
    Proposed LPF $\rightarrow$ Refinement &  7.1 $\rightarrow$  \underline{4.7} & \textbf{ 8.6} $\rightarrow$ \textbf{5.5}  &  16.4 $\rightarrow$ \underline{11.6} & \underline{12.3} $\rightarrow$ \underline{8.6} \\
    \bottomrule
    \end{tabular}
}
\end{table}

\begin{table}[ht]
\caption{
Quantitative results on the new PFAUST benchmark. 
The numbers correspond to the mean geodesic error (scaled by 100) without post-processing refinement. 
Best performance is marked in bold.
These results are consistent with those obtained on SHREC'16. 
As expected, the proposed method performs better than  previous unsupervised ones and achieves on par results as the supervised one. 
}
\label{tab:result_table pfaustrm}
\centering
%\resizebox{\textwidth}{!}{
    \begin{tabular}{c|c|c|c}
    \toprule
     \multicolumn{2}{c|}{} & \text{PFAUST-M} & \text{PFAUST-H} \\
     \midrule
     Supervised & DPFM \cite{DPFM} & 3.0 & 6.8 \\
     \midrule
     \multirow{3}{*}{Unsupervised} & Unsupervised-DPFM \cite{DPFM} & 9.3 & 12.7 \\
     & RobustFMnet \cite{cao2023unsupervised} & 7.9 & 12.4 \\
     & Proposed method & \textbf{5.1} & \textbf{7.9} \\
    \bottomrule
    \end{tabular}
%}
\end{table}

%_____________________________

\paragraph{Results.}
We present quantitative comparisons of the proposed method to the baseline methods in \cref{tab:result_table,tab:result_table pfaustrm} and qualitative ones in \cref{fig:qualitative_result_12}, see more examples in the supplementary.
On the CUTS dataset our method outperforms preexisting unsupervised approaches and almost bridges the gap between the performance of supervised and unsupervised methods for partial shape matching.
Our approach also performs state-of-the-art results on the HOLES benchmark, better than previous methods by a large margin. 
It even surpasses supervised methods such as DPFM.
This margin is consistent with our theoretical analysis regarding the error induced by the FM-layer on partial shape correspondence.
The reason is that shapes have larger missing regions in the HOLES  than in the CUTS dataset, and as shown in our analysis the error is proportional to the area of the missing parts. 
Furthermore, in the HOLES dataset, more points are closer to the carved-out boundaries than in the CUTS dataset. 
This complicates the task of the feature extractor to provide consistent features on the full and partial shapes. 
The reason is that many more points in the partial shape have significantly different local neighbourhood, whereas the feature extractor relies on the structure of such neighbourhoods to construct consistent point features.
Additionally, we show that the LPF approximation, which allows to avoid functional maps, achieves SOTA results in the unsupervised case of the CUTS dataset. 
For this approximation to hold, an important argument in its derivation  was that applying a partial shape low-pass filter to the low eigenfunctions of the full shape does not alter them by much \cref{eq:BPXtoY}. 
This fact holds when the boundaries are either short or close to the symmetry axes of the shapes, as discussed in \cite{pfm}. 
The LPF approximation is thus better tailored for datasets with such types of boundaries. 
The parts in the CUTS dataset exhibit such a behaviour, unlike those in the HOLES dataset.

The results on the PFAUST benchmark when compared to unsupervised approaches that are based on functional maps support our theoretical observations and demonstrate the power of the proposed method. 
As in the SHREC'16 HOLES dataset, the more holes there are, the better is the proposed approach, compared to alternative unsupervised methods.

%__________________________
\paragraph{Ablation study.}
The major difference between our approach and other methods is the order and type of operators we apply. 
Another significant change is the use of different loss functions, in particular $\mathcal{L}_{\text{G}}$, see Eq. (\ref{eq: L dist}). 
To evaluate the contribution of each of these design choices, we enrich our analysis by performing two types of ablation studies to the proposed method.
The first focuses on the loss functions, whereas the second is centred about the pipeline as a whole.
We perform the ablation studies on the well established SHREC'16 datasets.
\begin{table}[htbp]
\caption{
  Loss ablation study results. 
  The results demonstrate that the performance of our method mostly relies on $\mathcal{L}_\text{G}$, whereas $\mathcal{L}_{\text{orth}}$ and $\mathcal{L}_{\text{lpf}}$ act mostly as regularisation.
}
\label{tab: ablation losses}
\centering
{\footnotesize
%\resizebox{0.3\columnwidth}{!}{
    \begin{tabular}{c|c|c}
    \toprule
     Loss functions & \text{CUTS} & \text{HOLES}  \\
    \midrule
    $\mathcal{L}_\text{orth}$ & 24 & 14  \\
    $\mathcal{L}_\text{lpf}$ & 13 & 13  \\
     $\mathcal{L}_\text{G}$ & 10 & 14  \\
    $\mathcal{L}_\text{G} + \lambda_{\text{orth}}\mathcal{L}_\text{orth}$  & \textbf{6.9} & \textbf{11.4}  \\
    $\mathcal{L}_\text{G} + \lambda_{\text{lpf}}\mathcal{L}_\text{lpf}$  & 7.1 & 12.3  \\
    \bottomrule
    \end{tabular}
%}
}
\end{table}
%
%____
\paragraph{Loss analysis.}
In this experiment we deactivate, in turns, each part of the loss function and evaluate the outcome.
The results, see \cref{tab: ablation losses},  indicate that the Gromov part of the loss, Eq.~(\ref{eq: L dist}), is significant when operating on the CUTS benchmark, while the terms $\mathcal{L}_{\text{orth}}$ and  $\mathcal{L}_\text{lpf}$ serve mostly for regularization. 
As for the HOLES benchmark, the $\mathcal{L}_\text{G}$ part of the loss suffers from severe distortions of the geodesic distances that characterize that benchmark.
%_____________________
\begin{table}[ht]
\caption{
Pipeline ablation study results. 
The correspondence matrix is either estimated from the functional map extracted from a regularized FM-layer or, as we propose, directly from the cosine similarity between features. 
The results demonstrate the benefit of avoiding the use of an FM-layer in partial shape matching.
The effect is even more acute when considering more challenging scenarios, like the HOLES dataset.
}
\label{tab: pipeline ablation}
\centering
%\resizebox{0.33\columnwidth}{!}{
    \begin{tabular}{c|c|c}
    \toprule
     & \text{CUTS} & \text{HOLES} \\
    \midrule
    Regularized FM & 8.5 & 15.8 \\
    Proposed method & \textbf{6.9} & \textbf{11.4} \\
    \bottomrule
    \end{tabular}
%}
\end{table}
%
%__________________________________________
\paragraph{FM-layer.}
In this experiment we compare the proposed pipeline to a network that includes an FM-layer, where the refined features from the feature extractor serve as inputs to an FM-layer, which is regularized as done in recent functional map related publications, see e.g. \cite{fm_mask}.
This FM-layer applies a weighted mask based on the eigenvalues of the LBO to the functional map matrix. 
It was shown to give superior results than using the non-regularised version \cite{litany2017deep}. 

The loss $\mathcal{L}_\text{G}$ can be computed in the FM-layer pipeline using the correspondence matrix based on the estimated functional map ${\boldsymbol{P}_{yx}  = \boldsymbol \Psi_x \boldsymbol{C}_{yx} \boldsymbol \Phi^\top_y \boldsymbol{A}_y}$. 
Note, that both pipelines are trained using $\mathcal{L}_\text{G}$ and $\mathcal{L}_\text{orth}$, see Eq. (\ref{eq: TOTAL loss orth}), but not $\mathcal{L}_\text{lpf}$ as it is less suited for FM-layer based approaches.

The proposed approach outperforms ones with  FM-layers, see \cref{tab: pipeline ablation}, and performs exceptionally well on the HOLES benchmark. 
This aligns with our analysis as HOLES poses a harder challenge for the FM-layer than CUTS.

Note, that coupling the metric distortion Gromov-loss $\mathcal{L}_\text{G}$ with an FM-layer pipeline yields better results than previous approaches. 
It demonstrates that  $\mathcal{L}_G$
is better suited for partial shape matching than alternative traditional losses.

\section{Conclusion}
In this paper, we explored the problem of unsupervised partial shape matching. 
Our analysis unveiled a persistent obstacle when using spectral domains as part of neural network architectures rather than just as part of loss functions.
Our study showed that  using the functional maps framework via least-squares estimation inevitably introduces errors when shape partiality is involved, an issue that escalates with the increase in missing areas of the matched parts.

As a result of our findings, we proposed a novel approach that deviates from  conventional schools of thought. 
Rather than estimating the functional map on the computed features via least squares, and then extract the correspondences from it, we directly compute the correspondence between partial and full shapes through feature similarity.
The proposed method achieves superior results compared to previous methods on the SHREC'16 and the new PFAUST datasets, significantly surpassing  existing unsupervised methods in the field of partial shape matching of non-rigid shapes. 
The compelling results endorse the efficacy of our method and proves that spectral representations can be suited for partial shape correspondence when applied properly.

%{\color{blue}
The proposed approach obviously involves limitations. 
First, in its current form, it is not designed for part-to-part shape correspondence.
In that case, $\mathcal{L}_G$ needs to be revisited, since, when defining $\boldsymbol P$ we assume that each point in the partial shape has a corresponding point on the full shape.
Another issue is that geodesic distances can be significantly altered by cuts and holes, which affects the contribution of $\mathcal{L}_G$.
And yet, it is far more stable than current unsupervised losses based on functional maps, as demonstrated in the ablation study.
This issue is revisited in a follow-up paper \cite{bracha2024wormhole} where we introduce the {\it wormhole criterion} that filters out potentially altered geodesic distances. 
An additional limitation is that partial and full shapes should be of the same scale.
This is a common assumption in partial shape matching that is made by most methods, including those based on functional maps.
The missing parts also alter the LBO eigenfunctions, which affects DiffusionNet as it uses LBO eigendecomposition to learn heat diffusion on the shape. 
In practice, DiffusionNet mitigates this by learning to shorten the time scale in the presence of missing parts.
Finally, designing a scale-invariant approach, for example, by using a scale-invariant metric, or opting for anisotropic or asymmetric metrics \cite{weber2024finsler}, is of great interest that goes beyond the scope of this paper.

%}
%________

% ---- Bibliography ----
%
% BibTeX users should specify bibliography style 'splncs04'.
% References will then be sorted and formatted in the correct style.
%
\bibliographystyle{splncs04}
\bibliography{main}

\begin{thebibliography}{10}
\providecommand{\url}[1]{\texttt{#1}}
\providecommand{\urlprefix}{URL }
\providecommand{\doi}[1]{https://doi.org/#1}

\bibitem{aflalo2015optimality}
Aflalo, Y., Brezis, H., Kimmel, R.: On the optimality of shape and data representation in the spectral domain. SIAM Journal on Imaging Sciences  \textbf{8}(2),  1141--1160 (2015)

\bibitem{sgmds}
Aflalo, Y., Dubrovina, A., Kimmel, R.: Spectral generalized multi-dimensional scaling. International Journal of Computer Vision  \textbf{118},  380--392 (2016)

\bibitem{attaiki2023understanding}
Attaiki, S., Ovsjanikov, M.: Understanding and improving features learned in deep functional maps. In: Proceedings of the IEEE/CVF Conference on Computer Vision and Pattern Recognition. pp. 1316--1326 (2023)

\bibitem{DPFM}
Attaiki, S., Pai, G., Ovsjanikov, M.: {DPFM}: Deep partial functional maps. In: 2021 International Conference on 3D Vision (3DV). {IEEE} (Dec 2021)

\bibitem{bensaid2023partialhamiltonian}
Bensa{\"\i}d, D., Bracha, A., Kimmel, R.: Partial shape similarity by multi-metric hamiltonian spectra matching. In: International Conference on Scale Space and Variational Methods in Computer Vision. pp. 717--729. Springer (2023)

\bibitem{bensaid2023partialpiecewise}
Bensa{\"\i}d, D., Rotstein, N., Goldenstein, N., Kimmel, R.: Partial matching of nonrigid shapes by learning piecewise smooth functions. In: Computer Graphics Forum. vol.~42. Wiley Online Library (2023)

\bibitem{bracha2024wormhole}
Bracha, A., Dag{\`e}s, T., Kimmel, R.: Wormhole loss for partial shape matching. Advances in Neural Information Processing Systems  (2024)

\bibitem{bracha2020shape}
Bracha, A., Halim, O., Kimmel, R.: Shape correspondence by aligning scale-invariant {LBO} eigenfunctions. In: Eurographics Workshop on 3D Object Retrieval. The Eurographics Association (2020)

\bibitem{bronstein2009partial}
Bronstein, A.M., Bronstein, M.M., Bruckstein, A.M., Kimmel, R.: Partial similarity of objects, or how to compare a centaur to a horse. International Journal of Computer Vision  \textbf{84},  163--183 (2009)

\bibitem{bronstein2006efficient}
Bronstein, A.M., Bronstein, M.M., Kimmel, R.: Efficient computation of isometry-invariant distances between surfaces. SIAM Journal on Scientific Computing  \textbf{28}(5),  1812--1836 (2006)

\bibitem{bronstein2006generalized}
Bronstein, A.M., Bronstein, M.M., Kimmel, R.: Generalized multidimensional scaling: a framework for isometry-invariant partial surface matching. Proceedings of the National Academy of Sciences  \textbf{103}(5),  1168--1172 (2006)

\bibitem{TOSCA}
Bronstein, A.M., Bronstein, M.M., Kimmel, R.: Numerical geometry of non-rigid shapes. Springer Science \& Business Media (2008)

\bibitem{bronstein2010gromov}
Bronstein, A.M., Bronstein, M.M., Kimmel, R., Mahmoudi, M., Sapiro, G.: A {Gromov-Hausdorff} framework with diffusion geometry for topologically-robust non-rigid shape matching. International Journal of Computer Vision  \textbf{89}(2-3),  266--286 (2010)

\bibitem{cao2023unsupervised}
Cao, D., Roetzer, P., Bernard, F.: Unsupervised learning of robust spectral shape matching. ACM Transactions on Graphics (TOG)  (2023)

\bibitem{cosmo2016shrec}
Cosmo, L., Rodola, E., Bronstein, M.M., Torsello, A., Cremers, D., Sahillioglu, Y., et~al.: Shrec’16: Partial matching of deformable shapes. Proc. 3DOR  \textbf{2}(9), ~12 (2016)

\bibitem{geomfmnet}
Donati, N., Sharma, A., Ovsjanikov, M.: Deep geometric functional maps: Robust feature learning for shape correspondence. In: Proceedings of the IEEE/CVF Conference on Computer Vision and Pattern Recognition. pp. 8592--8601 (2020)

\bibitem{eisenberger2021neuromorph}
Eisenberger, M., Novotny, D., Kerchenbaum, G., Labatut, P., Neverova, N., Cremers, D., Vedaldi, A.: Neuromorph: Unsupervised shape interpolation and correspondence in one go. In: Proceedings of the IEEE/CVF Conference on Computer Vision and Pattern Recognition. pp. 7473--7483 (2021)

\bibitem{elad2003bending}
Elad, A., Kimmel, R.: On bending invariant signatures for surfaces. IEEE Transactions on pattern analysis and machine intelligence  \textbf{25}(10),  1285--1295 (2003)

\bibitem{ginzburg2020cyclic}
Ginzburg, D., Raviv, D.: Cyclic functional mapping: Self-supervised correspondence between non-isometric deformable shapes. In: Computer Vision--ECCV 2020: 16th European Conference, Glasgow, UK, August 23--28, 2020, Proceedings, Part V 16. pp. 36--52. Springer (2020)

\bibitem{halimi2019unsupervised}
Halimi, O., Litany, O., Rodola, E., Bronstein, A.M., Kimmel, R.: Unsupervised learning of dense shape correspondence. In: Proceedings of the IEEE Conference on Computer Vision and Pattern Recognition. pp. 4370--4379 (2019)

\bibitem{kim2011blended}
Kim, V.G., Lipman, Y., Funkhouser, T.: Blended intrinsic maps. ACM transactions on graphics (TOG)  \textbf{30}(4),  1--12 (2011)

\bibitem{fmm}
Kimmel, R., Sethian, J.A.: Computing geodesic paths on manifolds. Proceedings of the national academy of Sciences  \textbf{95}(15),  8431--8435 (1998)

\bibitem{kingma2014adam}
Kingma, D.P., Ba, J.: Adam: A method for stochastic optimization. arXiv preprint arXiv:1412.6980  (2014)

\bibitem{srfeat}
Li, L., Attaiki, S., Ovsjanikov, M.: {SRFeat}: Learning locally accurate and globally consistent non-rigid shape correspondence. In: International Conference on 3D Vision (3DV). IEEE (2022)

\bibitem{li2020shape}
Li, Q., Liu, S., Hu, L., Liu, X.: Shape correspondence using anisotropic chebyshev spectral cnns. In: Proceedings of the IEEE/CVF Conference on Computer Vision and Pattern Recognition. pp. 14658--14667 (2020)

\bibitem{litany2017deep}
Litany, O., Remez, T., Rodola, E., Bronstein, A., Bronstein, M.: Deep functional maps: Structured prediction for dense shape correspondence. In: Proceedings of the IEEE international conference on computer vision. pp. 5659--5667 (2017)

\bibitem{fsp}
Litany, O., Rodol{\`a}, E., Bronstein, A.M., Bronstein, M.M.: Fully spectral partial shape matching. In: Computer Graphics Forum. vol.~36, pp. 247--258. Wiley Online Library (2017)

\bibitem{loshchilov2016cosannealing}
Loshchilov, I., Hutter, F.: Sgdr: Stochastic gradient descent with warm restarts. In: International Conference on Learning Representations (2016)

\bibitem{marin2020correspondence}
Marin, R., Rakotosaona, M.J., Melzi, S., Ovsjanikov, M.: Correspondence learning via linearly-invariant embedding. Advances in Neural Information Processing Systems  \textbf{33},  1608--1620 (2020)

\bibitem{zoomout}
Melzi, S., Ren, J., Rodol{\`a}, E., Sharma, A., Wonka, P., Ovsjanikov, M.: Zoomout: spectral upsampling for efficient shape correspondence. ACM Transactions on Graphics (TOG)  \textbf{38}(6),  1--14 (2019)

\bibitem{memoli2012some}
M{\'e}moli, F.: Some properties of {Gromov--Hausdorff} distances. Discrete \& Computational Geometry  \textbf{48},  416--440 (2012)

\bibitem{memoli2005distance}
M{\'e}moli, F., Sapiro, G.: Distance functions and geodesics on submanifolds of $\mathbb{R}^d$ and point clouds. SIAM Journal on Applied Mathematics  \textbf{65}(4),  1227--1260 (2005)

\bibitem{memoli2005theoretical}
M{\'e}moli, F., Sapiro, G.: A theoretical and computational framework for isometry invariant recognition of point cloud data. Foundations of Computational Mathematics  \textbf{5},  313--347 (2005)

\bibitem{fm}
Ovsjanikov, M., Ben-Chen, M., Solomon, J., Butscher, A., Guibas, L.: Functional maps: a flexible representation of maps between shapes. ACM Transactions on Graphics (ToG)  \textbf{31}(4),  1--11 (2012)

\bibitem{pazi2020unsupervised}
Pazi, I., Ginzburg, D., Raviv, D.: Unsupervised scale-invariant multispectral shape matching. arXiv preprint arXiv:2012.10685  (2020)

\bibitem{rampini2019correspondence}
Rampini, A., Tallini, I., Ovsjanikov, M., Bronstein, A.M., Rodol{\`a}, E.: Correspondence-free region localization for partial shape similarity via hamiltonian spectrum alignment. In: 2019 International Conference on 3D Vision (3DV). pp. 37--46. IEEE (2019)

\bibitem{fm_mask}
Ren, J., Panine, M., Wonka, P., Ovsjanikov, M.: Structured regularization of functional map computations. In: Computer Graphics Forum. vol.~38, pp. 39--53. Wiley Online Library (2019)

\bibitem{faustRemeshed}
Ren, J., Poulenard, A., Wonka, P., Ovsjanikov, M.: Continuous and orientation-preserving correspondences via functional maps. ACM Transactions on Graphics (ToG)  \textbf{37}(6),  1--16 (2018)

\bibitem{pfm}
Rodol{\`a}, E., Cosmo, L., Bronstein, M.M., Torsello, A., Cremers, D.: Partial functional correspondence. In: Computer Graphics Forum. vol.~36, pp. 222--236. Wiley Online Library (2017)

\bibitem{roufosse2019unsupervised}
Roufosse, J.M., Sharma, A., Ovsjanikov, M.: Unsupervised deep learning for structured shape matching. In: Proceedings of the IEEE/CVF International Conference on Computer Vision. pp. 1617--1627 (2019)

\bibitem{sharma2020weakly}
Sharma, A., Ovsjanikov, M.: Weakly supervised deep functional maps for shape matching. Advances in Neural Information Processing Systems  \textbf{33},  19264--19275 (2020)

\bibitem{sharp2022diffusionnet}
Sharp, N., Attaiki, S., Crane, K., Ovsjanikov, M.: Diffusionnet: Discretization agnostic learning on surfaces. ACM Transactions on Graphics (TOG)  \textbf{41}(3),  1--16 (2022)

\bibitem{weber2024finsler}
Weber, S., Dag{\`e}s, T., Gao, M., Cremers, D.: Finsler-laplace-beltrami operators with application to shape analysis. In: Proceedings of the IEEE/CVF Conference on Computer Vision and Pattern Recognition. pp. 3131--3140 (2024)

\end{thebibliography}

% NOT OFFICIAL GUIDELINE
% Do not include supplementary in submitted or final pdf, create a separate file for that. 
% We put it here for easier cross references
\clearpage\newpage

\clearpage
\setcounter{page}{1}

{
\thispagestyle{empty}
\centering
\Large
\textbf{On Unsupervised Partial Shape Correspondence}\\
\vspace{0.5em}Supplementary Material \\
\vspace{1.0em}
}

%\maketitlesupplementary

% \appendix

\section{Proof of \cref{th: FM error continuous case}}
\label{sec: proof of FM error continuous case}
\begin{proof}
    By splitting 
    integrals    on
    $\mathcal{X}$ into 
    a
    sum of integrals over $\mathcal{Y}$ and $\mathcal{Z}$,
       we have,
    \begin{eqnarray}
        \hat{\boldsymbol{F}}_x &=& \langle \psi_{i}^{\mathcal{X}},f_{j}^{\mathcal{X}}\rangle_{\cal{X}}
                        \, =\,  \int_\mathcal{X} {\psi_{i}^{\mathcal{X}}}^*(x)f_{j}^{\mathcal{X}}(x)dx \cr
                        &= &\int_\mathcal{Y} {\psi_{i}^{\mathcal{X}}}^*(y)f_{j}^{\mathcal{X}}(y)dy +
                        \int_\mathcal{Z} {\psi_{i}^{\mathcal{X}}}^*(z)f_{j}^{\mathcal{X}}(z)dz \cr
                        &= &\langle \psi_{i}^{\mathcal{X}},f_{j}^{\mathcal{X}}\rangle_{\cal{Y}} +
                        \langle \psi_{i}^{\mathcal{X}},f_{j}^{\mathcal{X}}\rangle_{\cal{Z}} %\cr
                        \,= \, \hat{\boldsymbol {F}}^{(\mathcal{Y})} + \, \hat{\boldsymbol {F}}^{(\mathcal{Z})}.
                        \label{eq: continuous fx split in y and z}
    \end{eqnarray}
    The FM-layer computes the functional map in Eq. (\ref{eq: estimated FM pseudo inverse}). 
    We can plug in ${\hat{\boldsymbol F}}_x = \hat{\boldsymbol {F}}^{(\mathcal{Y})} + \hat{\boldsymbol {F}}^{(\mathcal{Z})}$ into Eq. (\ref{eq: estimated FM pseudo inverse}) and obtain
    $\hat{\boldsymbol{C}}_{yx} = \boldsymbol  C_{yx} + \boldsymbol C_{yx}^E$.
\end{proof}

\section{Proof of \cref{th: discrete case}}
\label{proof_discrete}
\begin{proof}
Let $n_x$, $n_y$, and $n_z$ be the number of vertices of $\mathcal{X}$, $\mathcal{Y}$, and $\mathcal{Z}$, respectively. 
Without loss of generality, we sort the vertices of $\mathcal{X}$  such that those corresponding to the vertices in $\mathcal{Y}$ appear first and in the same order as those of $\mathcal{Y}$. 
Thereby, we can write,
\begin{eqnarray}
   \boldsymbol  \Psi_x &=& \begin{bmatrix}
\boldsymbol \Psi^{(\mathcal{Y})} \\
\boldsymbol \Psi^{(\mathcal{Z})}
\end{bmatrix}
\label{psi_x},
\end{eqnarray}
where $\boldsymbol \Psi^{(\mathcal{Y})}\in\mathbb{R}^{n_y\times k}$ and $\boldsymbol \Psi^{(\mathcal{Z})}\in\mathbb{R}^{n_z\times k}$.

Denote, by $\boldsymbol F_x\in \mathbb{R}^{n_x\times d}$ and $\boldsymbol F_y\in \mathbb{R}^{n_y\times d}$  the feature matrices of $\mathcal{X}$ and $\mathcal{Y}$, respectively. 
Based on our ordering, we have that,
\begin{eqnarray}
   \boldsymbol F_x &=& \begin{bmatrix}
\boldsymbol F^{(\mathcal{Y})} \\
\boldsymbol F^{(\mathcal{Z})}
\end{bmatrix}.
\label{F_x}
\end{eqnarray}
Here, $\boldsymbol F^{(\mathcal{Y})}$ are the features extracted from vertices on $\mathcal{X}$ located on the subsurface $\mathcal{Y}$,
and similarly for $\boldsymbol F^{(\mathcal{Z})}$.
We can rewrite $\boldsymbol \Psi_x^\top \boldsymbol A_x \boldsymbol F_x$ using Eqs. (\ref{F_x}) and (\ref{psi_x}) as
\begin{eqnarray}
    \label{eq: discrete psix scal fx = y_part + z_part}
    \boldsymbol \Psi_x^\top \boldsymbol A_x \boldsymbol F_x & = & (\boldsymbol\Psi^{(\mathcal{Y})})^\top \boldsymbol A_y \boldsymbol F^{(\mathcal{Y})} + (\boldsymbol \Psi^{(\mathcal{Z})})^\top \boldsymbol A_z \boldsymbol F^{(\mathcal{Z})},
\end{eqnarray}
since vertices on both $\mathcal{X}$ and $\mathcal{Y}$ are associated to the same area on both shapes, meaning that  $\boldsymbol A_y = \boldsymbol A^{(\mathcal{Y})}$, and likewise ${\boldsymbol A_z = \boldsymbol A^{(\mathcal{Z})}}$,
where $\boldsymbol{A}^{(\mathcal{Y})} = \boldsymbol{A}_x|_\mathcal{Y}$ and $\boldsymbol{A}^{(\mathcal{Z})} =\boldsymbol{A}_x|_\mathcal{Z}$ are the restrictions of $\boldsymbol{A}_x$ to the matching part of $\mathcal{Y}$ and $\mathcal{Z}$ on $\mathcal{X}$, respectively.
Plugging Eq. (\ref{eq: discrete psix scal fx = y_part + z_part}) into Eq. (\ref{discrite_fm}), $ {\hat{\boldsymbol{C}}_{xy} = \boldsymbol C_{xy} + \boldsymbol C_{yx}^E}$  is the sum of two terms,
\begin{eqnarray}
    \boldsymbol  C_{yx}&=&  (\boldsymbol \Psi^{(\mathcal{Y})})^\top\boldsymbol A_y\boldsymbol F^{(\mathcal{Y})}\boldsymbol F_y^\top \boldsymbol A_y\boldsymbol \Psi_y\boldsymbol{Q}^{-1}_y,
    \label{C_tilde}
\end{eqnarray}
the self-functional map of $\mathcal{Y}$ with its correspondence in $\mathcal{X}$ with respect to the basis $\boldsymbol\Psi_x^{(\mathcal{Y})}$ and $\boldsymbol\Psi_y$,
and, 
\begin{eqnarray}
    \boldsymbol C^E_{yx} &=& (\boldsymbol \Psi^{(\mathcal{Z})})^\top\boldsymbol A_z\boldsymbol F^{(\mathcal{Z})}\boldsymbol F_y^\top \boldsymbol A_y\boldsymbol \Psi_y\boldsymbol{Q}^{-1}_y,
\end{eqnarray}
the unavoidable error injected into $C_{yx}$.
\end{proof}

%_____________________________ FFF
\section{Error analysis of the FM-layer}
\label{Error analysis of the FM-layer}
To better understand the two components that form $\hat{\boldsymbol{C}}_{yx}$, let us simplify the problem.
Assume that $\mathcal{X}$ is composed of two disconnected sub-surfaces $\mathcal{Y}$ and $\mathcal{Z}$. 
As the surfaces $\mathcal{Y}$ and $\mathcal{Z}$ are disconnected, the eigenfunctions of the LBO of $\mathcal{X}$ consist of two disjoint sets of functions, assuming different modes. 
These two disjoint sets are interleaving according to increasing eigenvalues, where one set contains functions that are the eigenfunction of the LBO of  $\mathcal{Y}$  extended to $\mathcal{Z}$ by taking zero values for every point on $\mathcal{Z}$, and vice versa for the other set.

Additionally, we assume that we have a feature extractor which is robust to partial shape matching, that is,
\begin{eqnarray}
    \boldsymbol{F}_y &=& \boldsymbol\Pi_{xy}^* \boldsymbol{F}_x,
\end{eqnarray}
where $\boldsymbol\Pi_{xy}^*$ is the exact correspondence matrix between $\mathcal{X}$ and $\mathcal{Y}$. 
Moreover, we assume that the extracted features have rank $d$.
We now look for the functional map $\boldsymbol C^*$ that would give us a perfect match between $\mathcal{X}$ and $\mathcal{Y}$. 
That is, 
\begin{eqnarray}
    \boldsymbol C^* &=& \boldsymbol\Psi_x^\top \boldsymbol A_x \boldsymbol\Pi_{yx}^* \boldsymbol\Psi_y ,
\end{eqnarray}
where, w.l.o.g. we choose $\boldsymbol{\Psi}_x$ and $\boldsymbol{\Psi}_y$ as the eigenfunctions of the LBO of surfaces $\mathcal{X}$ and $\mathcal{Y}$, respectively, and we sort the vertices of $\mathcal{X}$  such that those corresponding to the vertices in $\mathcal{Y}$ appear first and in the same order as those of $\mathcal{Y}$. 
we have that $\boldsymbol\Pi_{xy}^* = \boldsymbol J_{n_y}$ where 
$\boldsymbol J_{n_y} = 
\left[\begin{smallmatrix}\boldsymbol I_{n_y} \\ \boldsymbol{0} \end{smallmatrix}\right]
$ is the identity matrix until column $n_y$ and only zeros columns afterwards. 
Without loss of generality, we 
sort $\boldsymbol \Psi_x$ so that its leading eigenfunctions are those
related to $\mathcal{Y}$, and then those related to $\mathcal{Z}$.
Thus, we have,
\begin{eqnarray}
\boldsymbol C^* &=& \boldsymbol \Psi_x^\top \left[\begin{smallmatrix}\boldsymbol A_y \\ \boldsymbol{0}\end{smallmatrix}\right] \boldsymbol \Psi_y = 
\left[\begin{smallmatrix}\boldsymbol \Psi_y^\top \boldsymbol A_y \\ \boldsymbol{0}\end{smallmatrix}\right] \boldsymbol \Psi_y= \boldsymbol J_{n_y}.
\end{eqnarray}
Using our previous assumptions in Eq.~(\ref{c_yx})  gives us,
\begin{eqnarray}
\boldsymbol{C}_{yx} &=&  (\boldsymbol \Psi^{(\mathcal{Y})})^\top\boldsymbol A_y\boldsymbol F^{(\mathcal{Y})}\boldsymbol F_y^\top \boldsymbol A_y\boldsymbol \Psi_y\boldsymbol{Q}^{-1}_y \cr
&=& \left[\begin{smallmatrix}\boldsymbol \Psi_y^\top \cr
\boldsymbol{0}\end{smallmatrix}\right]\boldsymbol A_y\boldsymbol F_y\boldsymbol F_y^\top \boldsymbol A_y\boldsymbol \Psi_y\boldsymbol{Q}^{-1}_y\cr
&=& \left[\begin{smallmatrix}\boldsymbol \Psi_y^\top \boldsymbol A_y\boldsymbol F_y\boldsymbol F_y^\top \boldsymbol A_y\boldsymbol \Psi_y\boldsymbol{Q}^{-1}_y\cr
\boldsymbol{0}\end{smallmatrix}\right] 
\, =\,  \boldsymbol J_{n_y}.
\end{eqnarray}
This is the ideal functional map as it defined by the given mapping between  the surfaces $\mathcal{X}$ and $\mathcal{Y}$.
Therefore, the second term $\boldsymbol{C}^E_{yx}$ is a an error which is proportional to the area of $\mathcal{Z}$.

\section{The softmax operator compared to the FM-layer.}
\label{sec:The Softmax operator compared to the FM-Layer}

Pipelines using the FM-layer or ours using the Softmax operator all involve the feature product $\boldsymbol{F}^{(\mathcal{Z})}\boldsymbol{F}_y^\top$. 
Since the feature extractor is unaware of the locations of the missing parts, it cannot attribute trivial features, e.g. $\boldsymbol{0}$ only values, at the missing parts. 
Therefore, the term $\boldsymbol{F}^{(\mathcal{Z})}\boldsymbol{F}_y^\top$, present in both pipelines, introduces errors.
In the case of the FM-layer the dependence on this product is linear, whereas in our pipeline the error in the correspondence matrix estimated from the softmax operator is proportional to $\boldsymbol{E}^{-1}\exp(\boldsymbol{F}^{(\mathcal{Z})}\boldsymbol{F}_y^\top/\tau)$,
where 
$\boldsymbol{E} = \text{diag}(\sum_i \text{exp}((\boldsymbol{F}_x\boldsymbol{F}_y^\top/\tau)_{ij}))$. 
For any practical feature extractor, like DiffusionNet \cite{sharp2022diffusionnet}, there is usually good alignment of features of vertices on the partial shape and their counterpart on the full shape. Comparatively, features on the missing parts may fail to align as well. Thus, the normalisation $\boldsymbol{E}^{-1}$ reduces exponentially the influence of noisy correlations with missing parts. 
This is particularly the case for $\tau\rightarrow 0$ (as in our experiments), as then, the softmax with temperature hyperparameter $\tau$ collapses to the max operation. For example, consider three points $x$, $y$, and $z$ where $x\in\mathcal{X}\setminus\mathcal{Z}$, $y\in\mathcal{Y}$, and $z\in\mathcal{Z}$, having features $\boldsymbol{f}_x$, $\boldsymbol{f}_y$, and $\boldsymbol{f}_z$ respectively. If $\boldsymbol{f}_z \boldsymbol{f}_y^\top = 0.5$, $\boldsymbol{f}_x \boldsymbol{f}_y^\top = 0.7$, and $\tau=0.01$, then the error is close to $10^{-9}$ in the softmax operation, which is negligible to the error of $0.35$ induced in the FM-layer.

%_________________________________

\section{The PFAUST benchmark}
\begin{figure}[tbp]
  \centering
  % First image
  \includegraphics[width=0.5\columnwidth]{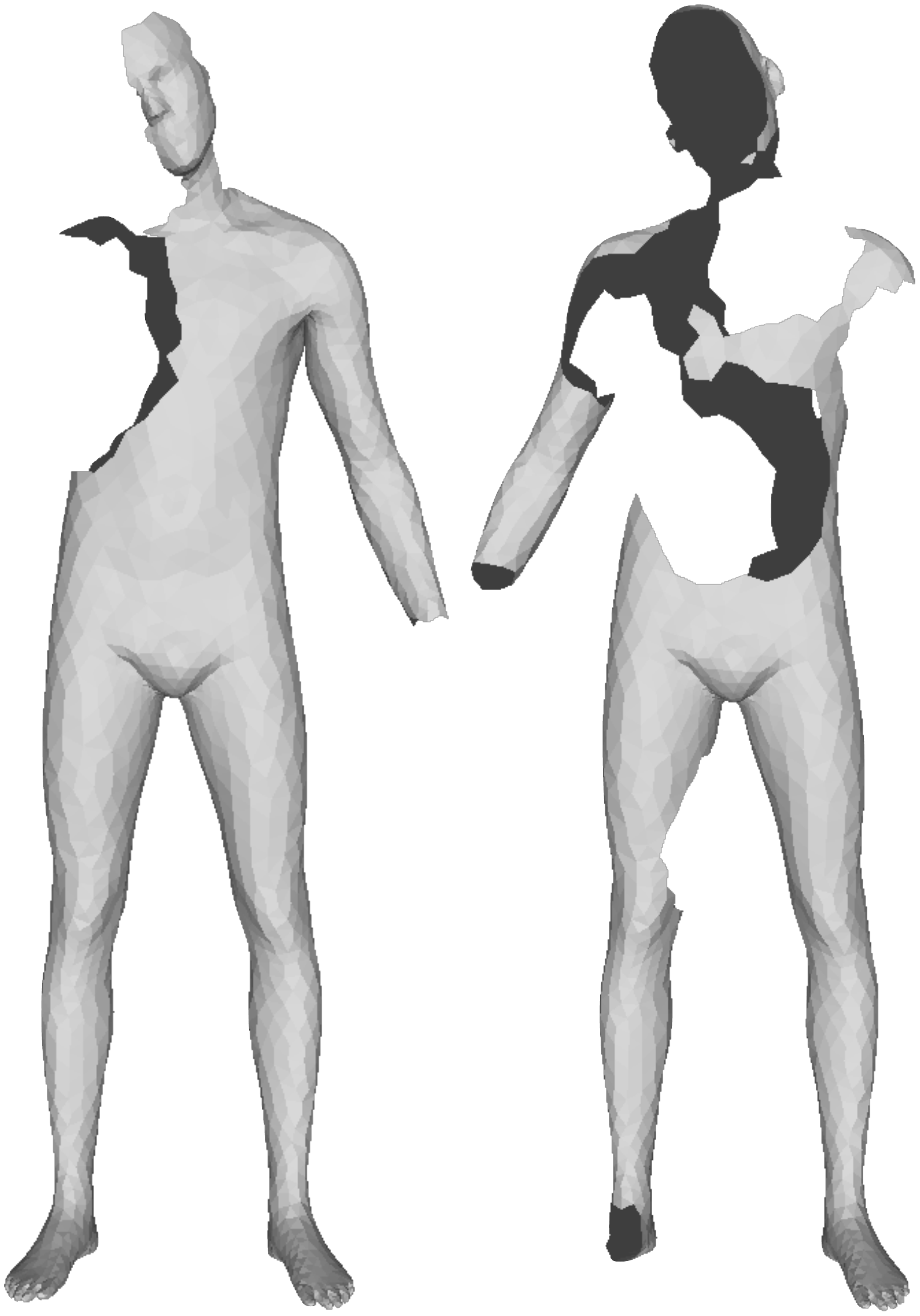}
  \caption{
  Example of shapes existing in our new PFAUST benchmark. The left shape is from PFAUST-M, while the right one is from PFAUST-H, which are of medium and hard difficulty, respectively.
  }
  % Second image, directly below the first
  \label{pfaust_example}
\end{figure}

We created the PFAUST benchmark from FAUST remeshed \cite{faustRemeshed} similarly to the way SHREC'16 \cite{cosmo2016shrec} was created from TOSCA \cite{TOSCA}. 
For each shape in FAUST, we randomly chose $m$ vertices, and for each selected vertex we removed all other vertices on the shape within a geodesic radius of $r$. 
As these holes may disconnect the shape, we then keep only  the connected component with the largest number of vertices. 
This process constructs the parts of our shapes. 
For the full shapes, we only chose subjects with ID ending with $0$. 
Shapes created from the train and test splits of FAUST remeshed populate our train and test splits respectively, and we provide for each partial shape its ground-truth correspondence matrix with its full shape.
Thus, the training set contains $8$ full shapes and $80$ partial shapes. 
The test set is analogous except for its number of shapes; $20$ partial and $2$ full shapes.
We created two datasets, PFAUST-M and PFAUST-H, generated with different choices of $m$ and $r$ to create two levels of missing regions.
A high number of small holes significantly changes the topology, which makes the partial shape correspondence task harder than with fewer albeit larger holes.
As such, PFAUST-M and PFAUST-H are of ``medium'' and ``hard'' difficulties, respectively.
We created PFAUST-M by taking $r = 0.16$ and $m = 4$, and PFAUST-H by choosing $r = 0.1$ and $m = 13$. 
See \cref{pfaust_example} for plots of example shapes from each dataset.

\section{Additional Implementation Considerations}

\paragraph{Our method.}
We trained our method using the Adam optimizer \cite{kingma2014adam}, with a learning rate of $10^{-3}$ and cosine annealing scheduler \cite{loshchilov2016cosannealing} with minimum learning rate parameter $\eta_{\text{min}}=10^{-4}$ and maximum temperature of $T_{\text{max}} = 300$ steps.
We train for 20000 iterations, and at test time our refinement process has 15 iterations.

\paragraph{On RobustFMnet.}
% %OPTION 1: MENTION TEST SET CONTAMINATION
The method by Cao et al. \cite{cao2023unsupervised} was claimed to produce state-of-the-art results when applied to various shape correspondence benchmarks including partial shape matching. 
And yet, the methodology proposed in \cite{cao2023unsupervised}, in spite of its conceptual beauty, involved difficulties in the evaluation phase. 
The results reported by Cao et al. were obtained by first pre-training  on the full shapes from the TOSCA dataset \cite{TOSCA}.
The problem is that the shapes and poses used in the TOSCA full shape dataset also play active part in the test set of SHREC'16.
The test set used by Cao et al. was thereby, unfortunately unintentionally, contaminated.
Recently, to avoid test set contamination, the authors pretrained the models on four external datasets.
However, the standard methodology in the field does not use any external datasets. As such, we did not include results from these updated models for fair comparison with all other methods.
It turns out, that without such pre-training, the method struggles to compete with previous approaches like \cite{pfm,fsp} for partial shape matching.

\begin{figure}[htbp]
    \centering 
    % {\resizebox{0.5\textwidth}{!}{
            \includegraphics[width=0.49\columnwidth]{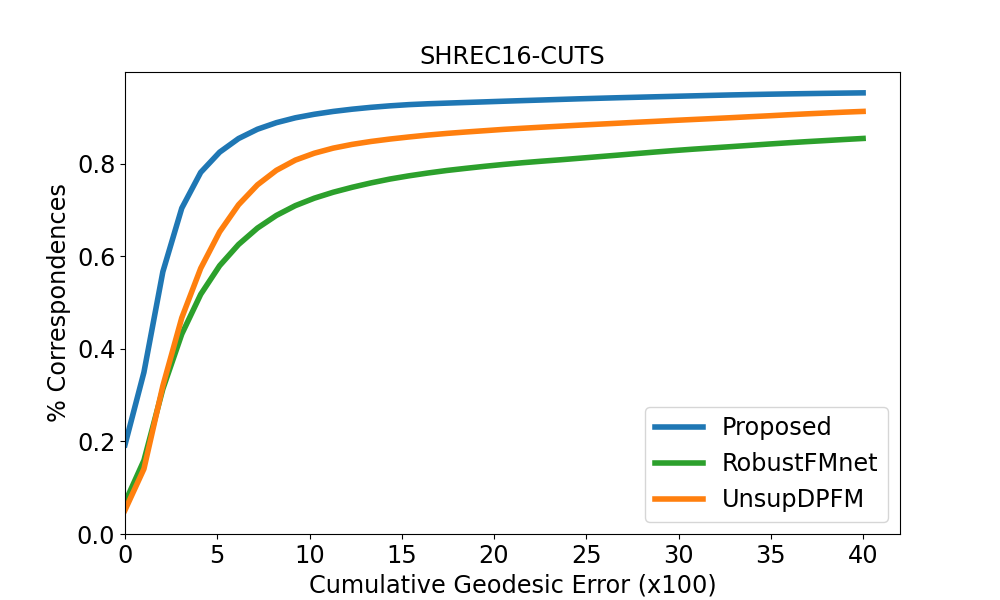}
            \includegraphics[width=0.49\columnwidth]{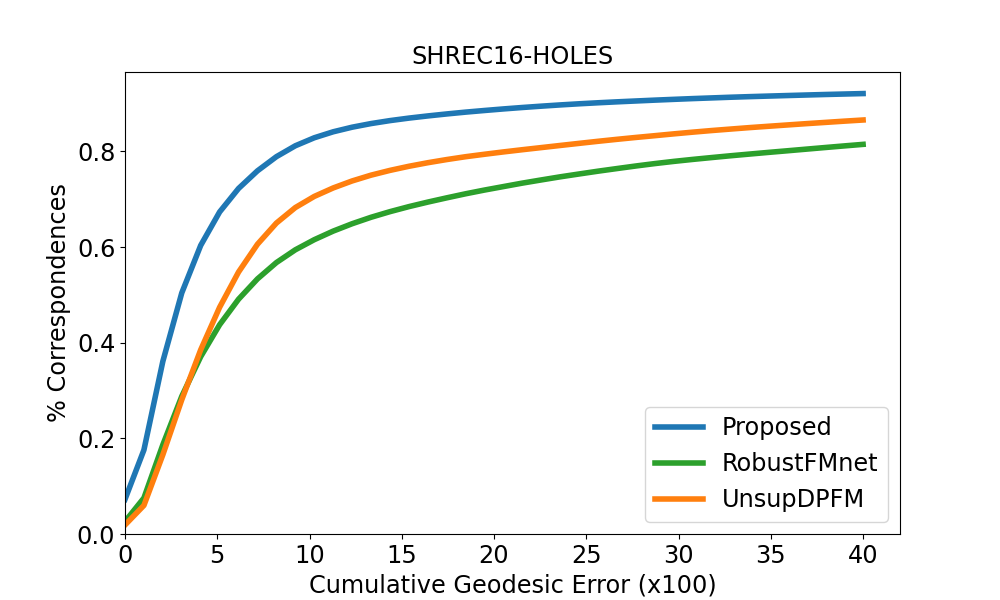}  
            \\
            \includegraphics[width=0.49\columnwidth]{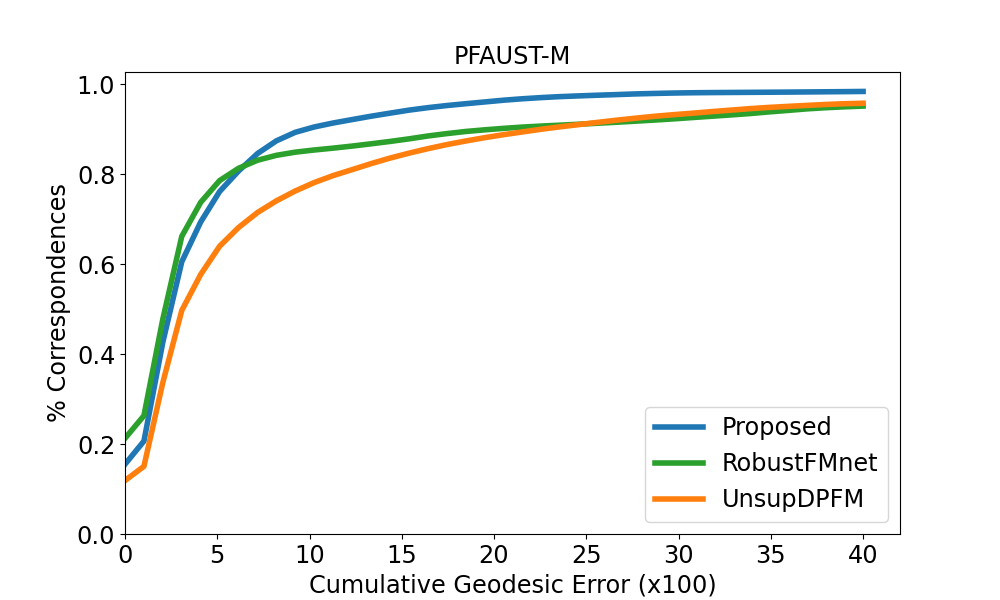}
            \includegraphics[width=0.49\columnwidth]{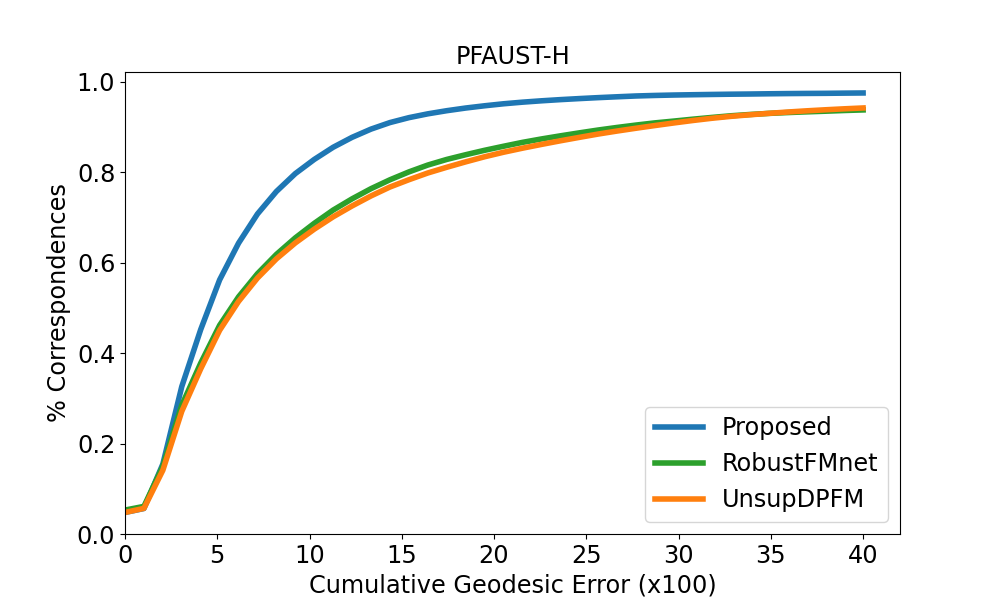}
     % }}
    \caption{PCK curves of existing unsupervised methods and ours on the test sets of SHREC'16 CUTS (top left), SHREC'16 HOLES (top right), PFAUST-M (bottom left) and PFAUST-H (bottom right). 
    Our method is systematically superior compared to competing unsupervised approaches.
    }
    \label{fig: pck curves shrec + pfaust}
\end{figure}

%_________________________________
\section{Further results}

\begin{figure*}[hbtb]
  \centering
  \includegraphics[width=\textwidth]{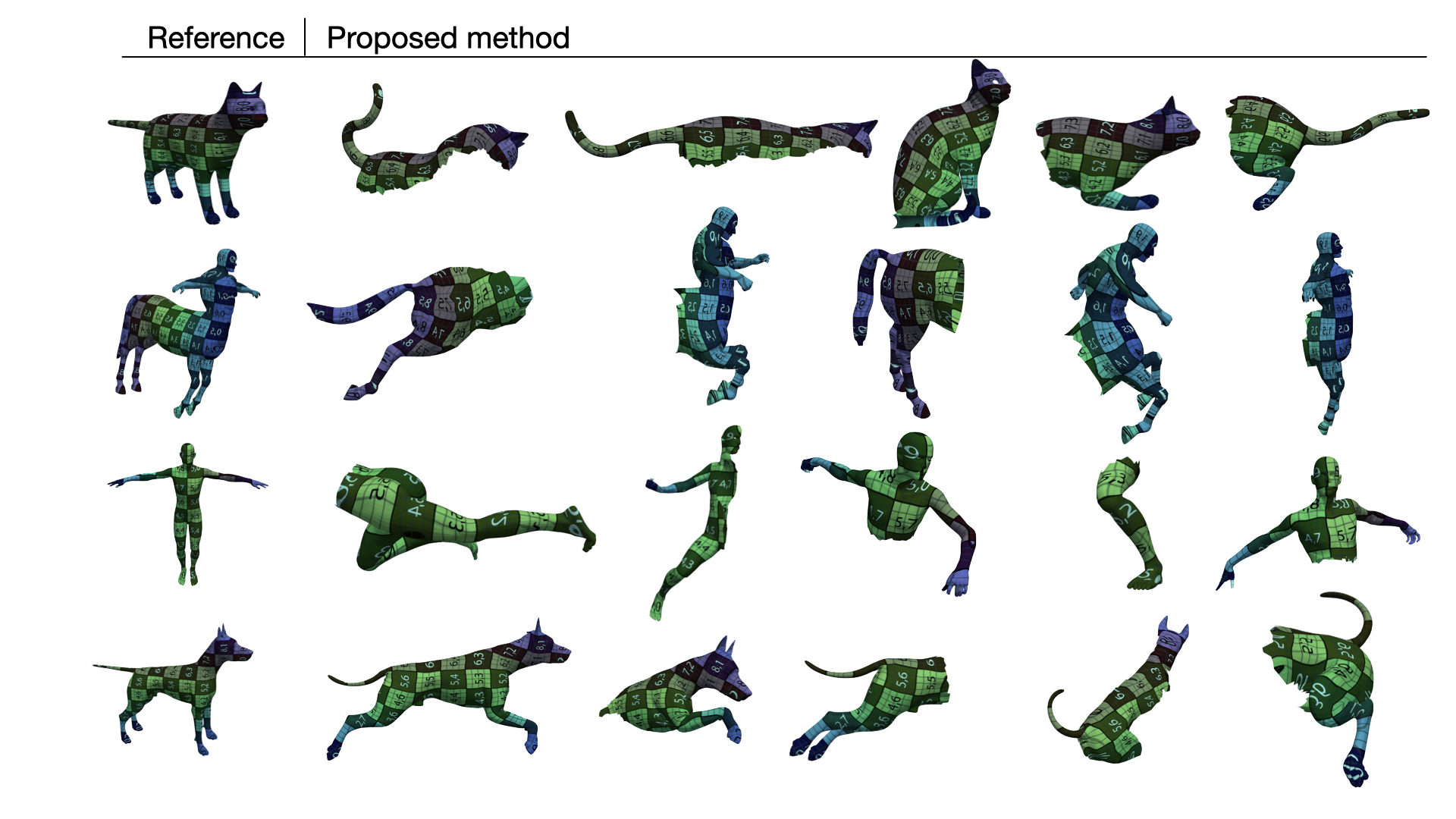}
  \includegraphics[width=\textwidth]{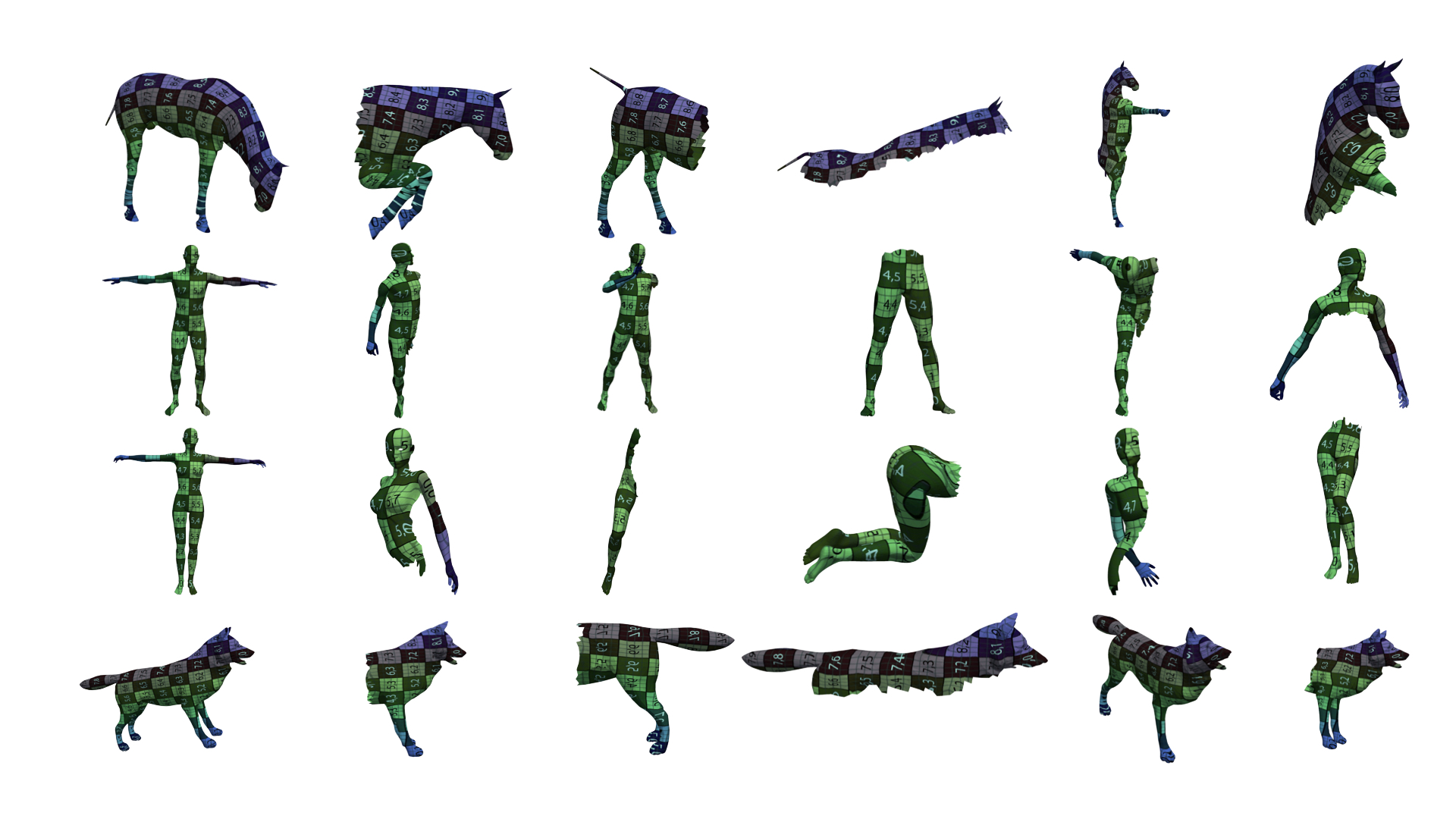} 
  \caption{Additional qualitative results on the SHREC'16 CUTS dataset. Zoom in for a better view.}
  \label{fig:qualitative_result supp mat shrec 16 cuts}
\end{figure*}

\begin{figure*}[htbp]
  \centering
  \includegraphics[width=\textwidth]{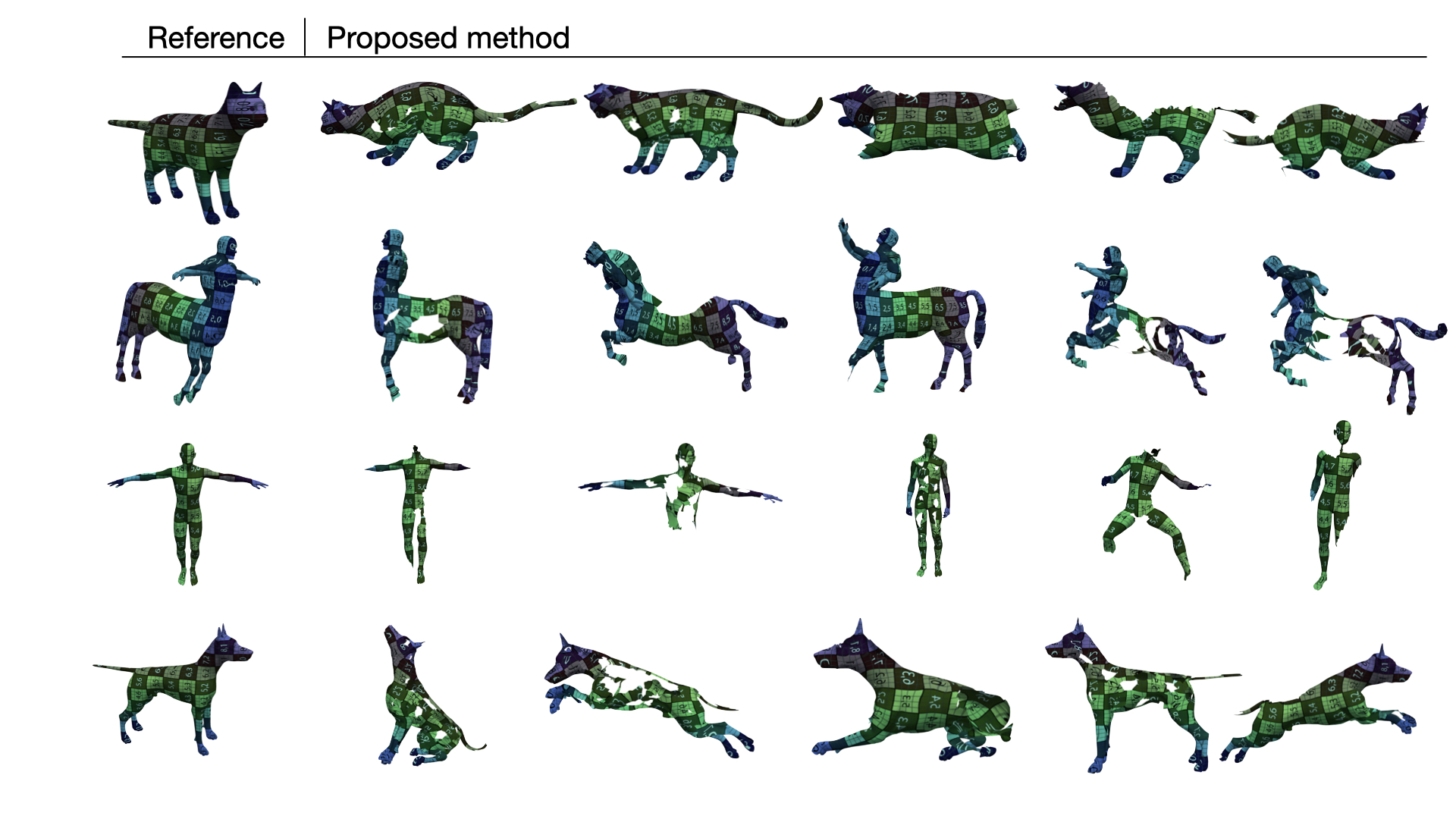}
  \includegraphics[width=\textwidth]{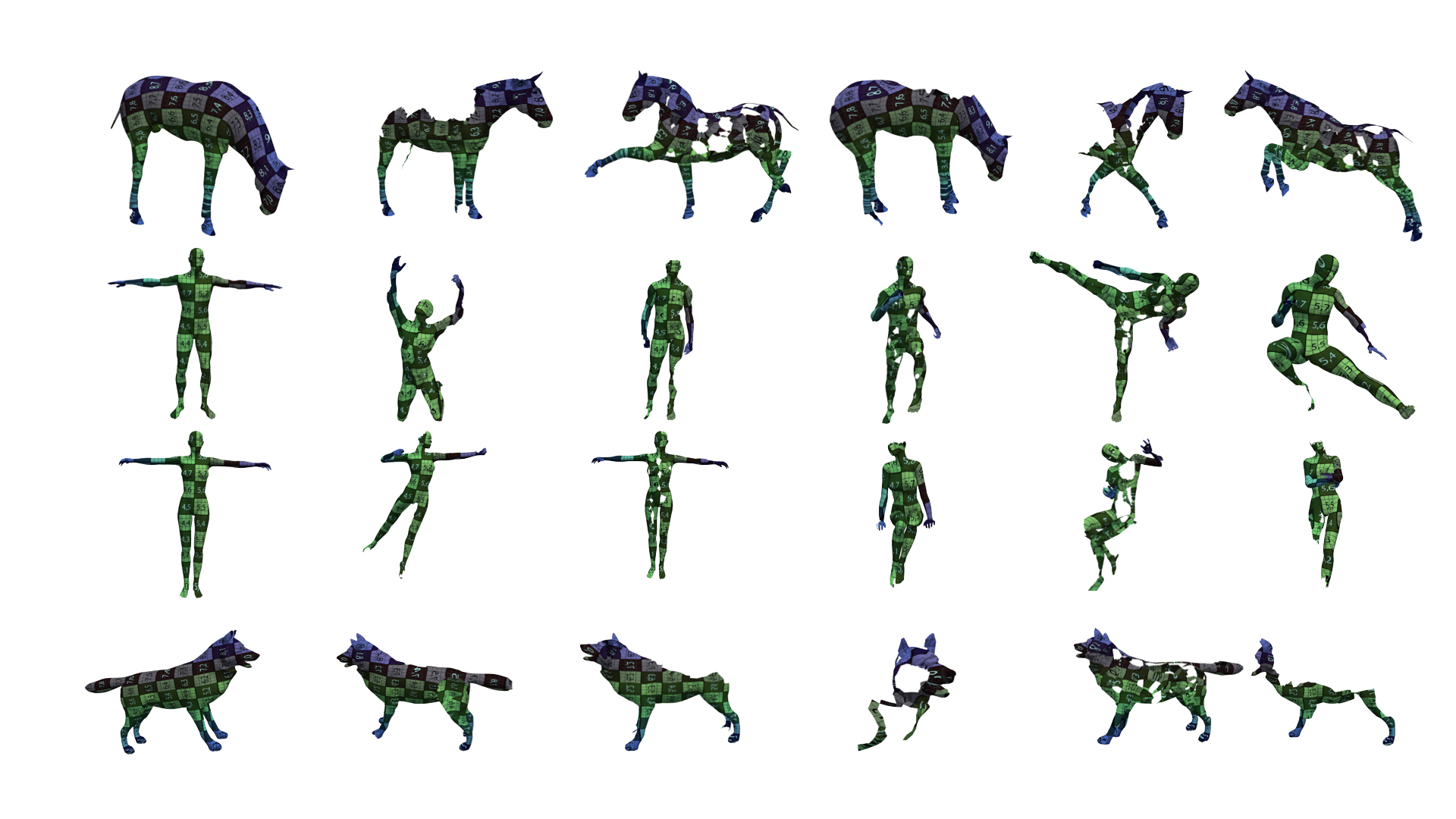} 
  \caption{Additional qualitative results on the SHREC'16 HOLES dataset. Zoom in for a better view.}
  \label{fig:qualitative_result supp mat shrec 16 holes}
\end{figure*}

\begin{figure*}[htbp]
  \centering
  \includegraphics[width=\textwidth]{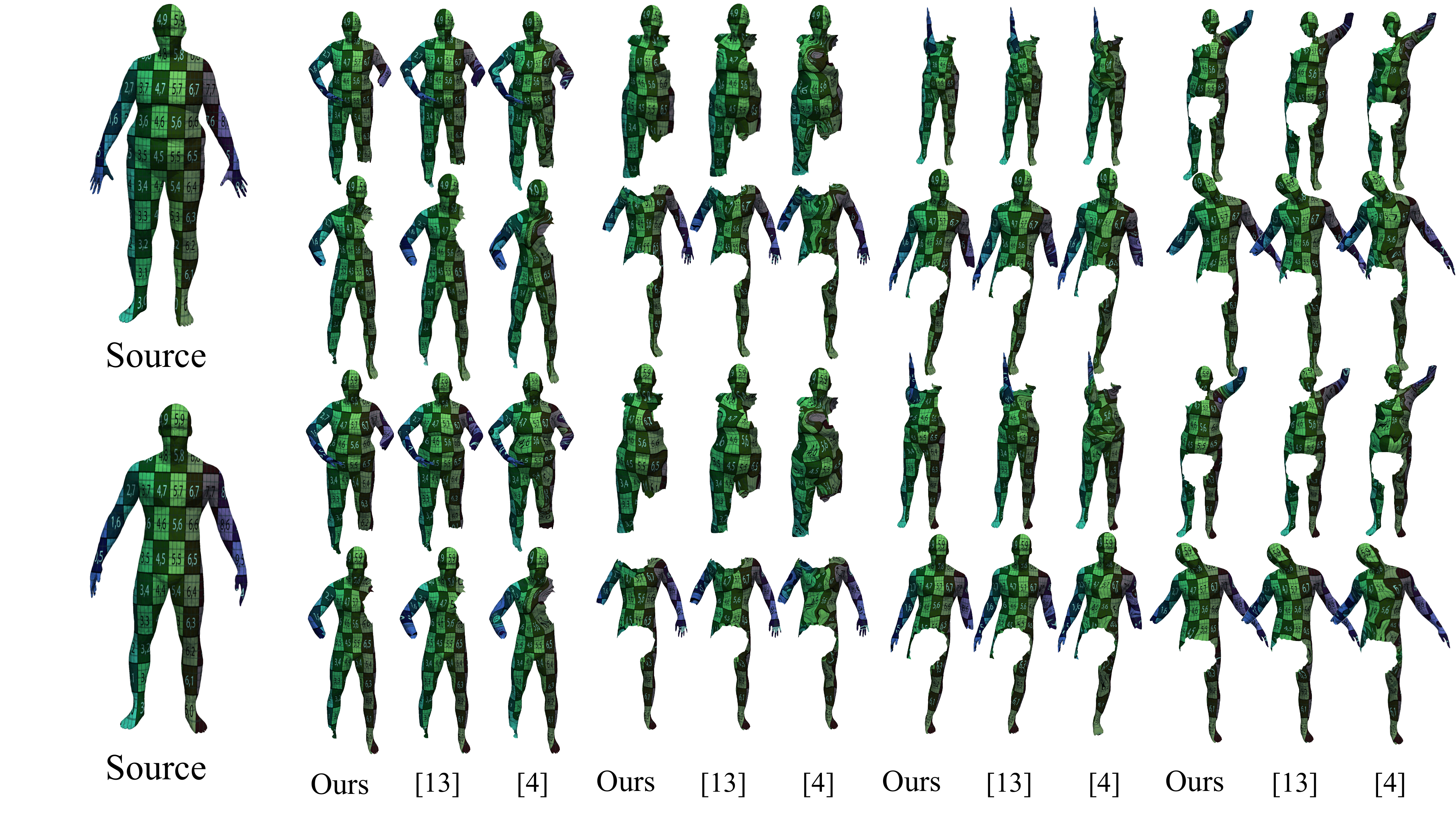}
  \caption{Qualitative results on PFAUST-M of our method and RobustFMnet \cite{cao2023unsupervised} and UnsupDPFM \cite{DPFM}, zoom in for a better view. 
  We obtain visually appealing results that outperform previous unsupervised methods. This figure also presents the shape partiality present in PFAUST-M, which mostly consists in body parts removal due to the size of the holes created on the original shapes.}
  \label{fig:qualitative_result supp mat pfaust m}
\end{figure*}

\begin{figure*}[tbp]
  \centering
  \includegraphics[width=\textwidth]{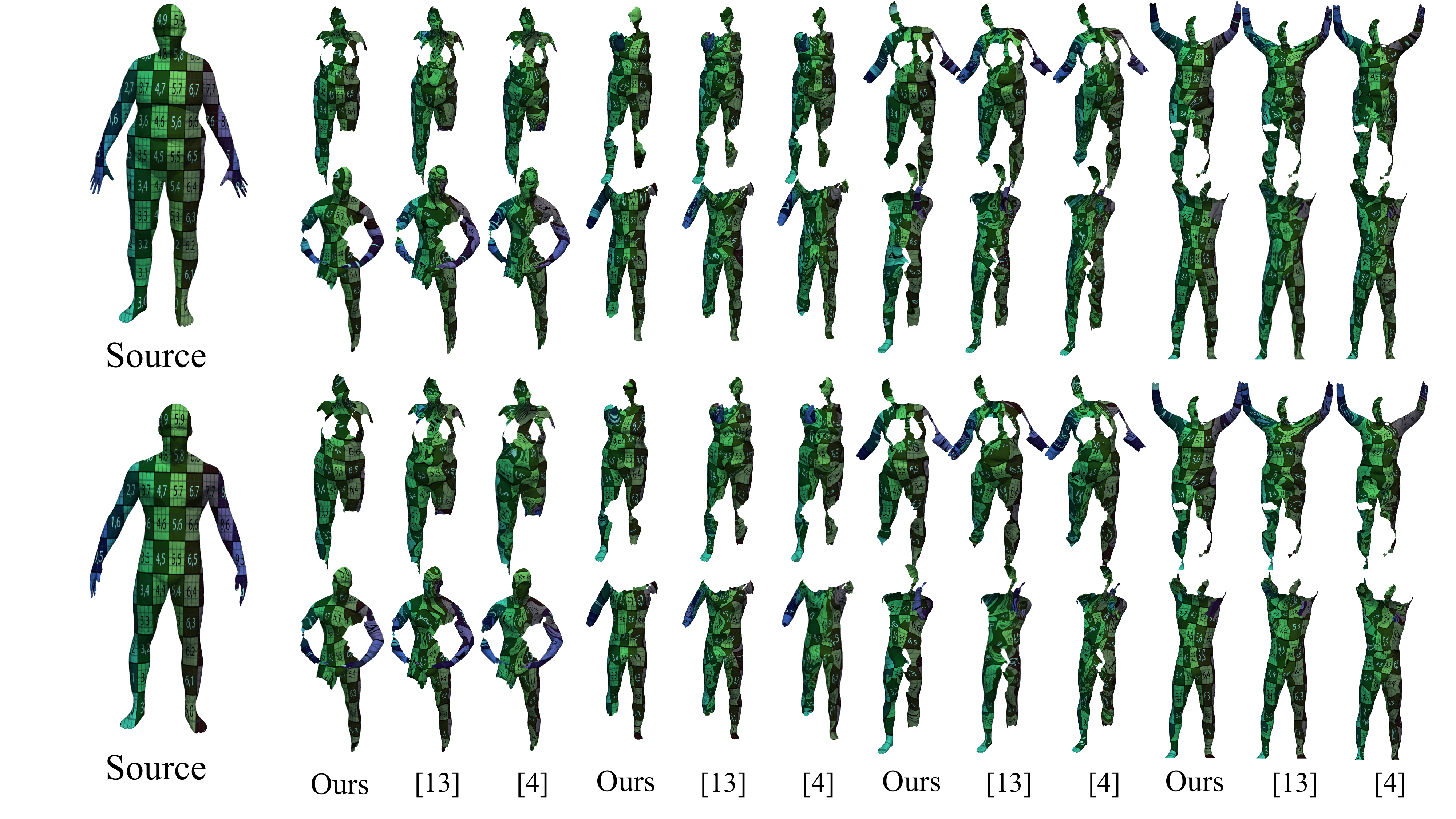}
  \caption{Qualitative results on PFAUST-H of our method and RobustFMnet \cite{cao2023unsupervised} and UnsupDPFM \cite{DPFM}. Zoom in for a better view. 
  We obtain visually appealing results that outperform previous unsupervised methods. This figure also presents the shape partiality present in PFAUST-H, which involves extremely challenging topology with $13$ holes.}
  \label{fig:qualitative_result supp mat pfaust h}
\end{figure*}

We provide the Percentage of Correct Keypoints (PCK) curves of ours and existing unsupervised methods, RobustFMnet \cite{cao2023unsupervised} and UnsupDPFM \cite{DPFM}, methods on the SHREC'16 and PFAUST benchmarks in \cref{fig: pck curves shrec + pfaust}.
We also provide additional qualitative results of our method for partial shape matching on the SHREC'16 CUTS (\cref{fig:qualitative_result supp mat shrec 16 cuts}) and HOLES (\cref{fig:qualitative_result supp mat shrec 16 holes}) datasets. 
Finally, we provide visual qualitative results on our new PFAUST-M (\cref{fig:qualitative_result supp mat pfaust m}) and PFAUST-H (\cref{fig:qualitative_result supp mat pfaust h}) datasets.

\end{document}